\documentclass[runningheads]{llncs}
\usepackage{graphicx}

\usepackage{tikz}
\usepackage{comment}
\usepackage{amsmath,amssymb}
\usepackage{color}
\usepackage{xcolor}
\usepackage{makecell}
\usepackage{multirow}
\usepackage{threeparttable}
\usepackage{subfig}
\usepackage{booktabs}
\newcommand\Tstrut{\rule{0pt}{2.2ex}}       

\usepackage{xspace}
\usepackage{textcomp}
\usepackage{gensymb} 
\usepackage{booktabs}
\usepackage{import}
\usepackage{multirow}
\usepackage{makecell}
\usepackage{bm}
\usepackage{capt-of}
\newcommand{\Tref}[1]{Table~\ref{#1}}
\newcommand{\eref}[1]{Eq.~\eqref{#1}}

\newcommand{\fref}[1]{Fig.~\ref{#1}}
\newcommand{\Fref}[1]{Figure~\ref{#1}}

\usepackage[pagebackref=true,breaklinks=true,colorlinks=true,bookmarks=false,linkcolor={red!50!black},urlcolor={darkgray!80!black},citecolor={green!50!black}]{hyperref}

\RequirePackage{enumitem}
\setlist[itemize]{nosep}
\usepackage[labelsep=period]{caption}
\renewcommand{\paragraph}[1]{\vspace{0.3em}\noindent \textbf{#1 \hspace{0.2em}}}
\newcommand{\etal}{\textit{et~al.}}
\newcommand{\eg}{\textit{e}.\textit{g}.}
\newcommand{\ie}{\textit{i}.\textit{e}.}

\newcommand{\mvdiligentdata}{DiLiGenT-MV benchmark\xspace}
\newcommand{\MVDiligentData}{DiLiGenT-MV Benchmark\xspace}
\newcommand{\syndataPS}{Synth$^\text{PS}$ dataset\xspace}
\newcommand{\syndataEnv}{Synth$^\text{Env}$ dataset\xspace}

\newcommand{\emphobj}[1]{\textit{#1}}

\begin{document}

\pagestyle{headings}
\mainmatter

\title{PS-NeRF: Neural Inverse Rendering for \\ Multi-view Photometric Stereo} 

\titlerunning{PS-NeRF: Neural Inverse Rendering for Multi-view Photometric Stereo}

\author{Wenqi Yang$^1$ \enspace Guanying Chen$^2$\thanks{Corresponding author} \enspace Chaofeng Chen$^3$  \\ Zhenfang Chen$^4$ \enspace Kwan-Yee K. Wong$^1$}

\authorrunning{W. Yang et al.}

\institute{$^1$The University of Hong Kong \quad $^2$FNii and SSE, CUHK-Shenzhen \\ $^3$Nanyang Technological University  \quad $^4$MIT-IBM Watson AI Lab}

\maketitle

\begin{abstract}
    Traditional multi-view photometric stereo (MVPS) methods are often composed of multiple disjoint stages, resulting in noticeable accumulated errors.
    In this paper, we present a neural inverse rendering method for MVPS based on implicit representation.
    Given multi-view images of a non-Lambertian object illuminated by multiple unknown directional lights, our method jointly estimates the geometry, materials, and lights.
    Our method first employs multi-light images to estimate per-view surface normal maps, which are used to regularize the normals derived from the neural radiance field.
    It then jointly optimizes the surface normals, spatially-varying BRDFs, and lights based on a shadow-aware differentiable rendering layer.
    After optimization, the reconstructed object can be used for novel-view rendering, relighting, and material editing.
    Experiments on both synthetic and real datasets demonstrate that our method achieves far more accurate shape reconstruction than existing MVPS and neural rendering methods.
    Our code and model can be found at \url{https://ywq.github.io/psnerf}.
\end{abstract}

\section{Introduction}
Multi-view stereo (MVS) is a technique for automated 3D scene reconstruction from a set of images captured from different viewpoints. 
As MVS methods rely on feature matching across different images, they generally assume the scene to be composed of textured Lambertian surfaces~\cite{seitz2006comparison,furukawa2009accurate,agarwal2011building} and their reconstructions often lack fine details.
Photometric stereo (PS), on the other hand, can recover per-pixel surface normals of a scene from single-view images captured under varying light directions. By utilizing shading information, PS method can recover fine surface details for both non-Lambertian and textureless objects~\cite{woodham1980ps,hayakawa1994photometric,shi2019benchmark}. However, single-view PS methods are not capable of recovering a full 3D shape.

To combine the merits of both techniques, multi-view photometric stereo (MVPS) methods are proposed to recover high-quality full 3D shapes for non-Lambertian and textureless objects~\cite{esteban2008multiview,park2016robust,li2020multi}.
Traditional MVPS methods are often composed of multiple disjoint stages~\cite{li2020multi,park2016robust}, leading to noticeable accumulated errors. 

\begin{figure}[t] \centering
    \includegraphics[width=\textwidth]{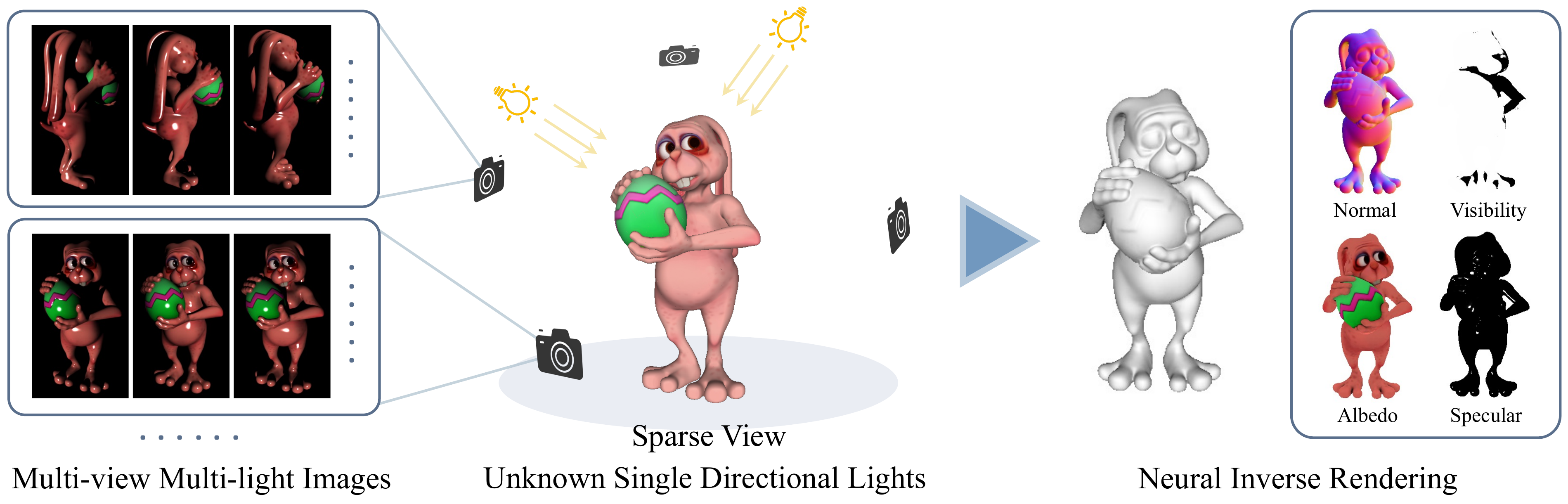}
    \caption{
    Our method takes multi-view multi-light images as input, and is able to reconstruct accurate surface and faithful BRDFs based on our shadow-aware renderer. Specifically, we only take images under \textbf{sparse} views with each view illuminated by multiple \textbf{unknown} single directional lights.
    } \label{fig:idea_illustration}
\end{figure}

Neural rendering methods have recently been introduced to tackle the problems of multi-view reconstruction and novel-view synthesis~\cite{yariv2020multiview,niemeyer2020differentiable,mildenhall2020_nerf_eccv20}. 
These methods work on multi-view images captured under fixed illuminations and show spectacular results. In addition to viewing directions, Kaya~\etal~\cite{kaya2022neural} use surface normals estimated from an observation map-based PS method to condition the neural radiance field (NeRF).
Although improved rendering results have been reported, this method has 4 fundamental limitations, namely 1. it requires calibrated lights as input to estimate per-view normal maps; 2. it takes surface normals as inputs for NeRF, making novel-view rendering difficult (if not impossible);
3. it does not recover BRDFs for the surface, making it not suitable for relighting; and 4. the PS network and NeRF are disjointed, and normal estimation errors from PS network will propagate to NeRF and cannot be eliminated.

To solve these challenges, we propose a neural inverse rendering method for multi-view photometric stereo (See \fref{fig:idea_illustration}).
Our method does not require calibrated lights. 
We first estimate per-view normal maps to constrain the gradient of density in NeRF. 
Surface normals, BRDFs, and lights are then jointly optimized based on a shadow-aware differentiable rendering layer.
By taking advantage of both multi-view and multi-light images, our method achieves far more accurate shape reconstruction results. Moreover, as our method explicitly models BRDFs and lights, it allows novel-view rendering, relighting, and material editing.

In summary, the key contributions of this paper are as follows:
\begin{itemize}
    \item[$\bullet$] We introduce a neural inverse rendering method for multi-view photometric stereo, which jointly optimizes shape, BRDFs, and lights based on a shadow-aware differentiable rendering layer.
    \item[$\bullet$] We propose to regularize the surface normals derived from the radiance field with normals estimated from multi-light images, which significantly improves surface reconstruction, especially for sparse input views (e.g., 5 views).
    \item[$\bullet$] Our method achieves state-of-the-art results in MVPS, and demonstrates that incorporating multi-light information appropriately can produce a far more accurate shape reconstruction.
\end{itemize}

\section{Related Work}

\paragraph{Single-view Photometric stereo (PS)}
Traditional PS methods rely on outlier rejection~\cite{wu2010robust,mukaigawa2007analysis,wu2010photometric}, reflectance model fitting~\cite{tozza2016direct,chung2008efficient,ikehata2014p}, or exemplars \cite{hertzmann2005example,hui2017shape} to deal with non-Lambertian surfaces.
Deep learning based PS methods solve this problem by learning the surface reflectance prior from a dataset~\cite{santo2017deep,chen2018ps,ikehata2018cnn,Taniai18,chen2020deepps,logothetis2021px}. 
These methods typically train a deep network to learn the mapping from image measurements to surface normals.
Recently, many efforts have been devoted to reduce the number of images required~\cite{li2019learning,zheng2019spline} and to estimate light directions in an uncalibrated setup~\cite{chen2019self,kaya2021uncalibrated,lu2018symps}.

\paragraph{Multi-view Photometric Stereo (MVPS)}
MVPS combines the advantages of MVS and PS methods, leading to more accurate surface reconstructions.
Traditional MVPS methods assume a simplified surface reflectance~\cite{lim2005passive,esteban2008multiview,wu2010fusing,park2016robust,logothetis2019differential}. They first apply MVS to reconstruct a coarse shape from multi-view images and adopt PS to obtain per-view surface normals from multi-light images. They then refine the coarse shape using the obtained normals.
Zhou~\etal~\cite{li2020multi,zhou2013multi} propose a method to deal with isotropic materials. Their method 
first reconstructs sparse 3D points from multi-view images and identifies per-view iso-depth contours using multi-light images. It then recovers a complete 3D shape by propagating the sparse 3D points along the iso-depth contours.
Kaya~\etal~\cite{kaya2022uncertainty} propose an uncertainty-aware deep learning based method to integrate MVS depth maps with PS normal maps to produce a full shape.
Another branch of methods adopts a co-located camera-light setup for joint reflectance and shape recovery~\cite{cheng2021multi,li2018materials,nam2018practical,wang2020non}.

\paragraph{Neural Rendering}
Neural rendering methods have achieved great successes in novel-view synthesis and multi-view reconstruction~\cite{tewari2021advances,sitzmann2019scene,yariv2020multiview,niemeyer2020differentiable}.
In particular, neural radiance field (NeRF)~\cite{mildenhall2020_nerf_eccv20} achieves photo-realistic view synthesis by representing a continuous space with an MLP which maps 5D coordinates (\ie, 3D point and view direction) to density and color.
Many follow-up works are proposed to improve the reconstructed shape~\cite{oechsle2021unisurf,wang2021neus}, rendering speed~\cite{liu2020_nsvf_nips20,Reiser2021_kiloNeRF_iccv21,garbin2021_fastnerf_arxiv}, and robustness~\cite{martin2021_nerfw_cvpr21,Zhang20arxiv_nerf++,barron2021mip}.
However, these methods essentially treat surface points as light sources, and thus cannot disentangle materials and lights~\cite{wood2000surface}.

Some methods have been proposed to jointly recover shape, materials, and lights~\cite{zhang2021physg,boss2021nerd}.
PhySG~\cite{zhang2021physg} and NeRD~\cite{boss2021nerd} adopt Spherical Gaussian (SG) representation for BRDFs and environment illumination to enable fast rendering. 
Neural-PIL~\cite{boss2021neural} replaces the costly illumination integral operation with a simple network query. 
These methods~\cite{zhang2021physg,boss2021nerd,boss2021neural}, however, ignore cast shadows during optimization.
NeRV~\cite{srinivasan2021nerv} explicitly models shadow and indirect illumination, but it requires a known environment map.
NeRFactor~\cite{zhang2021nerfactor} proposes a learned auto-encoder to represent BRDFs and pre-extracts a light visibility buffer using the obtained mesh.
NRF~\cite{bi2020neural} optimizes a neural reflectance field assuming a co-located camera-light setup.

Similar to our proposed method, Kaya~\etal~\cite{kaya2022neural} introduce a NeRF-based method for MVPS.
Their method~\cite{kaya2022neural} assumes calibrated light directions and takes the estimated normals as input to condition their NeRF. It thus cannot disentangle surface materials and light directions.
\Tref{tab:relatedwork} summarizes the differences between our method and existing neural inverse rendering methods. Our method is the only one that explicitly models surface reflectances and lights under a multi-view photometric stereo setup.

\begin{table}[t] \centering
    \caption{Comparisons among different neural inverse rendering methods (MVI stands for multi-view images).} \label{tab:relatedwork}
    
\resizebox{\textwidth}{!}{
\begin{tabular}{l|*{4}{c}c}
    \toprule
    \makebox[0.18\textwidth][l]{\textbf{Method}} & \makebox[0.25\textwidth]{\textbf{Input}} & \makebox[0.1\textwidth]{\textbf{Shape}} & \makebox[0.28\textwidth]{\textbf{BRDF}} & \makebox[0.25\textwidth]{\textbf{Lighting}} & \textbf{Shadow} \\[0pt]
    \hline
    NeRV~\cite{srinivasan2021nerv} & MVI (Fixed lighting) & Density & Microfacet model & Known Envmap & Yes \\[0pt]
    PhySG~\cite{zhang2021physg} & MVI (Fixed lighting) & SDF & Microfacet (SGs) & Unknown Envmap (SGs) & No \\[0pt]
    NeRFactor~\cite{zhang2021nerfactor} & MVI (Fixed lighting) & Density & Learned BRDF & Unknown Envmap & Yes \\[0pt]
    NeRD~\cite{boss2021neural} & MVI (Varying lighting) & Density & Microfacet (SGs) & Unknown Envmap (SGs) & No \\[0pt]
        NRF~\cite{bi2020neural} & MVI (Co-located light) & Density & Microfacet model & Known Co-located light & Yes \\[0pt]
    \hline
    KB22~\cite{kaya2022neural} & MVI (Multi-light) & Density & No & No & No \\
    Ours & MVI (Multi-light) & Density & Mixture of SGs & Unknown Multi-light & Yes \\[0pt]
    \bottomrule
\end{tabular}
}

\end{table}


\newcommand{\viewdir}{\boldsymbol{d}}
\newcommand{\pointcolor}{\boldsymbol{c}}
\newcommand{\density}{\sigma}
\newcommand{\albedo}{\rho_d}
\newcommand{\rough}{\rho_s}
\newcommand{\envmap}{\boldsymbol{L}_{SG}}
\newcommand{\ray}{\boldsymbol{r}}
\newcommand{\pixelcolor}{\boldsymbol{C}}
\newcommand{\coarse}{{(c)}}
\newcommand{\refine}{{(r)}}
\newcommand{\loss}{\mathcal{L}}
\newcommand{\gt}{\mathrm{(gt)}}
\newcommand{\uncertain}{{(\tau)}}
\newcommand{\point}{\boldsymbol{x}}
\newcommand{\img}[2]{I_{#1}^{#2}}
\newcommand{\nview}{M}
\newcommand{\viewindx}{N_p}
\newcommand{\nlight}{L_m}
\newcommand{\imgset}{\mathcal{I}}
\newcommand{\camloc}{\boldsymbol{o}}

\newcommand*\diff{\mathop{}\!\mathrm{d}}
\newcommand*\Diff[1]{\mathop{}\!\mathrm{d^#1}}
\newcommand{\lout}{\hat{I}}
\newcommand{\lin}{L_i}
\newcommand{\din}{\boldsymbol{w}_i}
\newcommand{\dout}{\boldsymbol{w}_o}
\newcommand{\normal}{\boldsymbol{n}}
\newcommand{\vis}{n}
\newcommand{\brdf}{f_r}
\newcommand{\nmlp}{f_n}
\newcommand{\vmlp}{f_v}
\newcommand{\surf}{\mathcal{S}}
\newcommand{\surfpts}{\mathcal{S}_{\boldsymbol{x}}}
\newcommand{\surfnorm}{\mathcal{N}_{\sigma}}
\newcommand{\surfvis}{\mathcal{V}_{\sigma}}
\newcommand{\psnormalmap}{\mathcal{N}_{m}}
\newcommand{\R}{\mathcal{R}}


\newcommand{\stageone}{geometry modeling from photometric images}
\newcommand{\stagetwo}{Occlusion-aware High-fidelity Neural Inverse Rendering}

\section{Methodology}

\subsection{Overview}
\begin{figure}[t] \centering
    \includegraphics[width=\textwidth]{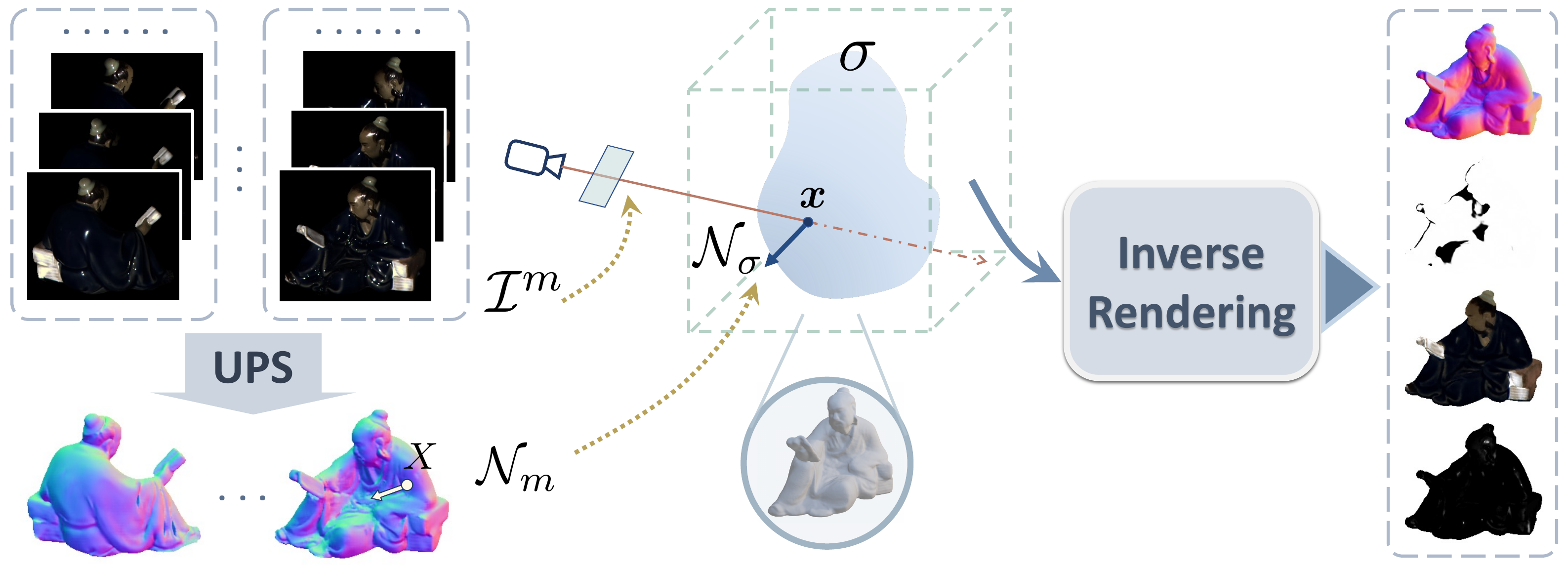}
    \caption{Given multi-view multi-light images, we first obtain guidance normal maps via uncalibrated photometric stereo (UPS) to regularize the neural density field, which encourages accurate surface reconstruction. We then perform neural inverse rendering to jointly optimize surface normals, BRDFs and lights based on the initial shape.} \label{fig:network_arch}
\end{figure}

Given multi-view and multi-light images\footnote{\ie, multiple images are captured for each view, where each image is illuminated by a single unknown directional light.} of an object taken from $\nview$ sparse views, our goal is to simultaneously reconstruct its shape, materials, and lights.
We denote the MVPS image set as $\imgset$, and multi-light images for each view $m$ as $\imgset^m = \{\img{1}{m},\img{2}{m},\dots, \img{\nlight}{m}\}$.
Note that the number of lights for each view can vary.
\Fref{fig:network_arch} illustrates the overall pipeline of our method.

Inspired by the recent success of neural radiance field~\cite{mildenhall2020_nerf_eccv20} for 3D scene representation, we propose to represent the object shape with a density field.
Our method consists of two stages to make full use of multi-view multi-light images. 

In the first stage, we estimate a guidance normal map $\psnormalmap$ for each view, which is used to supervise the normals derived from the density field. This direct normal supervision is expected to provide a strong regularization on the density field, leading to an accurate surface.
In the second stage, based on the learned density field as the shape prior, we jointly optimize the surface normals, materials, and lights using a shadow-aware rendering layer.  

\subsection{Stage I: Initial Shape Modeling}

In the first stage, we optimize a neural radiance field with surface normal regularizations to represent the object shape.

\paragraph{Neural Radiance Field}
NeRF employs an MLP to map a 3D point $\point_k$ in space and view direction $\viewdir$ to density $\density_k$ and color $\pointcolor_k$, \ie,
\begin{align}
    \label{eq:nerfmlp}
    (\density_k, \pointcolor_k) & = \mathrm{MLP}_{\mathrm{NeRF}} (\point_k, \viewdir).
\end{align}

Given a pixel in an image, its color can be computed by integrating the color of the points sampled along its visual ray $\ray$  through volume rendering, \ie,
\begin{align}
    \label{eq:volumerender}
    \tilde{\pixelcolor}(\ray) = \sum_{k=1}^K T_k(1 - \exp(-\density_k\delta_k))\pointcolor_k, \quad
    T_k  = \exp\left(-\sum_{j=1}^{k-1} \density_j\delta_j\right),
\end{align}
where $\delta_{k}=t_{k+1}-t_{k}$ is the distance between adjacent sampled points. 

Typically, a NeRF is fitted to a scene by minimizing the reconstruction error between the rendered image $\pixelcolor$ and the input image $\img{}{}$, \ie,
\begin{align}
    \loss'_c &= \sum \| \pixelcolor - \img{}{}\|_2^2.
\end{align}
However, as the surface geometry has no direct supervision, the density field is generally very noisy.
Recently, UNISURF~\cite{oechsle2021unisurf} improves the surface quality of NeRF by gradually shrinking the sampling range of a ray for volume rendering, leading to a smoother surface. 
However, as shown in our experiments, the shape recovered  by UNISURF~\cite{oechsle2021unisurf} is still not satisfactory. 

\paragraph{Surface Normal Regularization} The above observation motivates us to introduce regularizations for the surface geometry. 
Notably, state-of-the-art uncalibrated photometric stereo (UPS) method, such as \cite{chen2019self}, can estimate a good normal map from single-view multi-light images. We therefore use the normal map $\psnormalmap$ estimated by \cite{chen2019self} to regularize the density field by minimizing
\begin{align}
    \loss'_n= \sum \| \surfnorm - \mathcal{T}_{m2w}({\psnormalmap})\|_2^2,
    \quad \surfnorm \left(\point\right)=\frac{\nabla \density\left(\point\right)}{\left\|\nabla \density\left(\point\right)\right\|_{2}}, \label{eq:surf_normal}
\end{align}
where $\surfnorm$ is the surface normal derived from the gradient of the density filed $\density$ based on the expected depth location, and $\mathcal{T}_{m2w}$ is transformation that transforms the view-centric normals from the camera coordinate system to the world coordinate system. We also include the normal smoothness regularization~\cite{oechsle2021unisurf} with $\epsilon \sim \mathcal{N}(0, 0.01)$, \ie,
\begin{align}
    \label{eq:normal_smooth}
    \loss'_{ns} = \sum \| \surfnorm(\point) - \surfnorm(\point + \epsilon)\|_2^2.
\end{align}

We adopt UNISURF~\cite{oechsle2021unisurf} as our radiance field method for initial shape modeling, and the overall loss function for Stage~I is given by
\begin{align}
    \loss' &=  \alpha_1\loss'_c + \alpha_2\loss'_n +\alpha_3\loss'_{ns},
\end{align}
where $\alpha_*$ denotes the loss weights.

\begin{figure}[t] \centering
    \includegraphics[width=\textwidth]{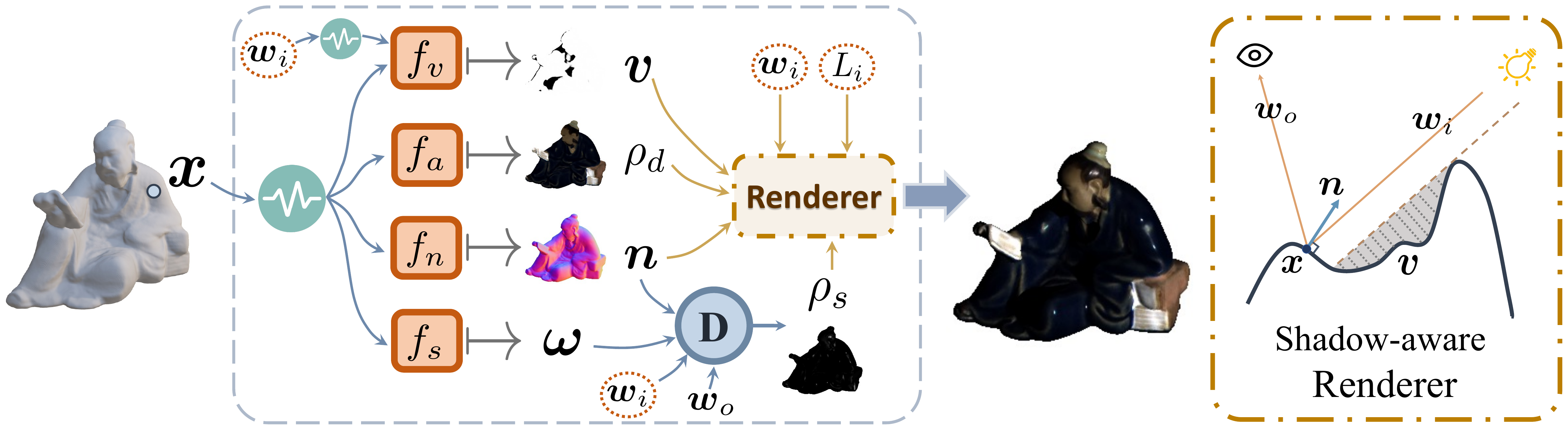}
    \caption{In Stage II, we model normals, BRDFs, and light visibility of the scene with MLPs. The weights of the MLPs and lights are jointly optimized to fit the input images.} \label{fig:stage2_arch}
\end{figure}

\subsection{Stage II: Joint Optimization with Inverse Rendering}

With the initial shape (\ie, $\density$ and $\surfnorm$) from Stage~I, we are able to jointly optimize the surface normals, spatially-varying BRDFs, and lights based on a shadow-aware rendering layer (see~\fref{fig:stage2_arch}). Specifically, we first extract the surface from the density field via root-finding \cite{niemeyer2020differentiable,mescheder2019occupancy} similar to \cite{oechsle2021unisurf}.
We then model surface normals, BRDFs, and light visibility of the scene with MLPs.
The weights of the MLPs and the lights are then jointly optimized to fit the input multi-view and multi-light images. In the following subsections, we will describe the formulation of our rendering layer and each component in detail.

\paragraph{Rendering Equation}
The rendering equation for a non-Lambertian surface point $\point$ can be written as~\cite{kajiya1986rendering}
\begin{align}
    \label{eq:render_eq_general}
    \lout (\dout; \point) = \int_\Omega \lin (\din) \brdf (\dout, \din; \point) (\din \cdot \normal) \diff \din,
\end{align}
where $\din$ and $\dout$ are the incident light direction and view direction respectively, $\brdf (\dout, \din; \point)$ represents a general BRDF at location $\point$. 
$\lin (\din)$ is the light intensity along $\din$ and $\lout (\dout; \point)$ is the integrated radiance over the upper-hemisphere $\Omega$.

By assuming a directional light and considering light visibility, the rendering equation can be rewritten as
\begin{align}
    \label{eq:render_eq_ours}
    \lout (\dout,\din; \point) = \vmlp (\din; \point) \lin (\din) \brdf (\dout, \din; \point) (\din \cdot \normal),
\end{align}
where $\vmlp (\din; \point)$ indicates the visibility of light along $\din$ at $\point$, and models cast-shadow in the rendered image.

\paragraph{Shape Modeling}
In Stage~I, we optimize a radiance field with surface normal regularizations to produce an initial density field $\density$. Note the normals used to regularize $\surfnorm$ are estimated by a PS method~\cite{chen2019self} which inevitably contain estimation errors. As a result, the derived normals $\surfnorm$ might not be accurate in some regions.
To refine the normals, we use an MLP $\nmlp(\point) \mapsto \normal $ to model the distribution of surface normals. This MLP will be optimized by an image fitting loss.
To encourage the refined normals to not deviate too much from the derived normals, we use the derived normals to regularize the output of $\nmlp(\point)$ by minimizing
\begin{align}
    \loss''_{n} = \sum \left\| \nmlp(\point) - \surfnorm(\point) \right\|_2^2.
\end{align}

\paragraph{Visibility Modeling}
Given the density field $\density$, a surface point $\point$, and a light direction $\din$, the light visibility $\surfvis(\point, \din)$ can be computed by applying volume rendering to calculate the accumulated density along the ray directed from $\point$ to the light source~\cite{zhang2021nerfactor}.

Since we perform ray-marching to calculate visibility for each point and each query light direction, it will be time-consuming to compute light visibility for an environment map lighting. Besides, the computed visibility might also be noisy.
We therefore model the distribution of light visibility using another MLP $\vmlp(\point, \din) \mapsto v$, which is regularized by the computed visibility by minimizing
\begin{align}
    \loss''_{v} = \left\| \vmlp(\point, \din) - \surfvis(\point, \din) \right\|_2^2.
\end{align}

\paragraph{Material Modeling}
As in previous works~\cite{zhang2021physg,boss2021nerd,zhang2021nerfactor}, we assume the BRDF model can be decomposed into diffuse color $\albedo$ and specular reflectance $\rough$, \ie, $\brdf(\point,\din,\dout) = \albedo + \rough (\point,\din,\dout)$.
For diffuse color, we use an MLP $f_a(\point) \mapsto \albedo$ to model the albedo for a surface point $\point$.

For specular component, one may adopt a reflectance model (\eg, Microfacet) to model the specular reflectance and estimate its parameters (\eg, roughness)~\cite{srinivasan2021nerv,li2018learning}.
However, we found it difficult to model the specular effects of real-world objects by directly estimating the roughness parameter.
Instead, we propose to fit the specular reflectance with a weighted combination of specular basis following~\cite{hui2017shape,li2022neural}.

We assume isotropic materials and simplify the input to a half-vector $\boldsymbol{h}$ and normal $\normal$ according to \cite{rusinkiewicz1998new}, and define a set of Sphere Gaussian (SG) basis as
\begin{align}
    D(\boldsymbol{h}, \boldsymbol{n}) = G(\boldsymbol{h}, \boldsymbol{n} ; \lambda)=\left[e^{\lambda_{1}\left(\boldsymbol{h}^{T} \boldsymbol{n}-1\right)}, \cdots, e^{\lambda_{k}\left(\boldsymbol{h}^{T} \boldsymbol{n}-1\right)}\right]^{T},
\end{align}
where $\lambda_* \in \R_+$ denotes specular sharpness.
We introduce an MLP $f_s(\point) \mapsto \boldsymbol{\omega}$ to model the spatially-varying SG weights, and specular reflectance can then be recovered as
\begin{align}
    \rough = \boldsymbol{\omega}^T D(\boldsymbol{h}, \boldsymbol{n}).
\end{align}

To encourage a smooth albedo and specular reflectance distribution, we impose smoothness losses $\loss_{as}^R$ and $\loss_{ss}^R$ (defined similarity as in \eref{eq:normal_smooth}) for $f_a(\point)$ and $f_s(\point)$ respectively.

\paragraph{Light Modeling}
Each image is illuminated by a directional light, which is parameterized by a 3-vector light direction and a scalar light intensity.
We set the light directions and intensities as learnable parameters, and initialize them by the lights estimated by the UPS method~\cite{chen2019self}. The light parameters will be refined after the joint optimization. 

\paragraph{Joint Optimization}
Based on our scene representation, we can rerender the input images using the differentiable rendering equation.
Given multi-view and multi-light images, 
we optimize the normal, visibility, and BRDF MLPs together with the light parameters to minimize the image reconstruction loss, given by
\begin{align}
    \loss''_c = \sum \| \hat{I} - \img{}{}\|_2^2,
\end{align}
where $\hat{I}$ is the rerendered image and $I$ is the corresponding input image.

The overall loss function used for our neural inverse rendering stage is
\begin{align}
    \loss'' = \beta_1 \loss''_c + \beta_2\loss''_n  +\beta_3\loss''_v + \beta_4 \loss''_{as} +\beta_5 \loss''_{ss},
\end{align}
where $\beta_*$ denotes the corresponding loss weight.

\section{Experiments}

\subsection{Implementation details}
Please refer to supplementary materials for implementation details. 
We first use all $96$ lights for training and evaluation, and then demonstrate that our framework can support arbitrary sparse light sources in Sec.~\ref{sec:analysis_nlight}.

\paragraph{Evaluation Metrics} We adopt commonly used quantitative metrics for different outputs. Specifically, we use mean angular error (MAE) in degree for surface normal evaluation under test views, and Chamfer distance for mesh evaluation\footnote{We rescale the meshes into the range of $[-1,1]$ for all the objects, and show Chamfer distance in the unit of $mm$).}. 
Following \cite{oechsle2021unisurf}, we extract meshes using the MISE algorithm~\cite{mescheder2019occupancy}. 
PSNR, SSIM~\cite{wang2004image}, and LPIPS~\cite{zhang2018unreasonable} are used to evaluate the reconstructed images.

\subsection{Dataset}
\paragraph{Real Data}
We adopted the widely used \mvdiligentdata~\cite{li2020multi} for evaluation. It consists of $5$ objects with diverse shapes and materials. Each object contains images captured from $20$ views. For each view, $96$ images are captured under varying light directions and intensities. Ground-truth meshes are provided. 
In our experiments, we sample $5$ testing views with equal interval, and take the rest $15$ views for training.
Note that our method assumes unknown light direction and light intensity in evaluation.

\paragraph{Synthetic Data}
To enable more comprehensive analysis, we rendered a synthetic dataset with $2$ objects (\ie, \emphobj{BUNNY},  \emphobj{ARMADILLO}) with Mitsuba~\cite{jakob2010mitsuba}.
We rendered objects under two sets of lightings, one with directional lights, denoted as \syndataPS, and the other with environment map, denoted as \syndataEnv.  We randomly sampled $20$ camera views on the upper hemisphere, where $15$ views for training and $5$ views for testing. 
We used directional lights in the same distribution as \mvdiligentdata for each view.
For the synthetic dataset, we set the same light intensity for each light source.


\subsection{Comparison with MVPS Methods}

We compared our method with state-of-the-art MVPS methods~\cite{li2020multi,park2016robust} on \mvdiligentdata, where all $20$ views are used for optimization. It should be noted that our method does not use calibrated lights as~\cite{li2020multi,park2016robust}.

\begin{table}[t] \centering
    \caption{Results of different MVPS methods on \mvdiligentdata.}
    \label{tab:ps_shape}

\resizebox{\textwidth}{!}{
\begin{tabular}{l|*{5}{c}|c|*{5}{c}|c}
    \toprule
    & \multicolumn{6}{c|}{Chamfer Dist$\downarrow$ } &  \multicolumn{6}{c}{Normal MAE$\downarrow$} \\
    Method 
    & \emph{BEAR}  & \emph{BUDDHA}   & \emph{COW}   & \emph{POT2}   & \emph{READING}  & \emph{Average} 
     & \emph{BEAR}  & \emph{BUDDHA}   & \emph{COW}   & \emph{POT2}   & \emph{READING}  & \emph{Average} 
    \\
    \hline
    PJ16~\cite{park2016robust} 
    &  19.58  & 11.77  & \textbf{9.25}  & 24.82  &  22.62  & 17.61
    & 12.78  & 14.68  & 13.21  & 15.53  & 12.92  & 13.83 
    \\
    LZ20~\cite{li2020multi} 
    &  8.91 &  13.29 &  14.01 & 7.40  &  24.78 & 13.68
    & 4.39  & 11.45  & \textbf{4.14}  & 6.70  & 8.73  & 7.08   
    \\
    Ours 
    & \textbf{8.65} & \textbf{8.61} & 10.21 & \textbf{6.11} & \textbf{12.35} & \textbf{9.19}
    & \textbf{3.54}  & \textbf{10.87}  & 4.42  & \textbf{5.93}  & \textbf{8.42}  & \textbf{6.64} 
    \\
    \bottomrule
\end{tabular}
}
\end{table}
\begin{figure}[t] \centering
    \vspace{-1em}
    \includegraphics[width=0.95\textwidth]{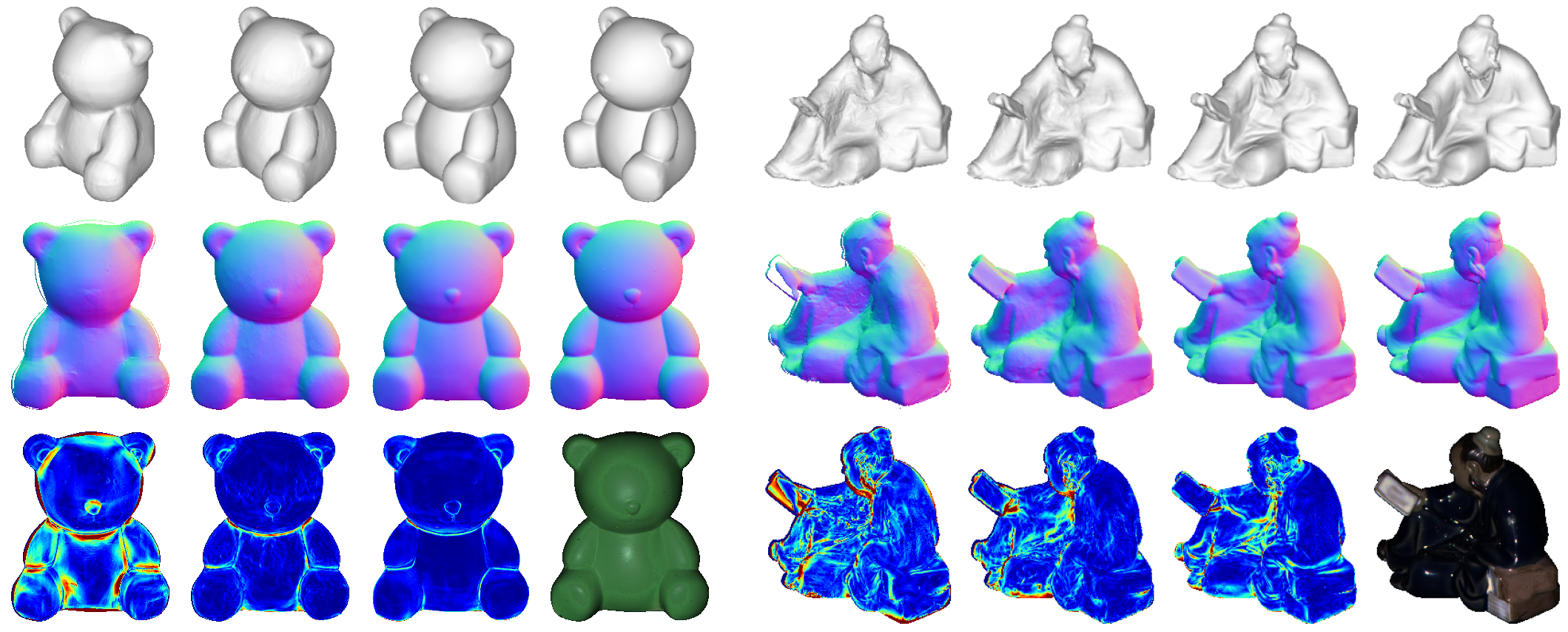}
    \includegraphics[width=0.035\textwidth]{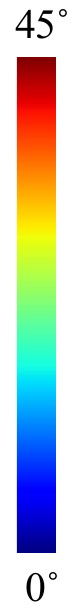}\\
    \vspace{-1em}
    \subfloat{\resizebox{\textwidth}{!}{
    \begin{tabular}{*{4}{>{\centering\arraybackslash}p{1.6cm}}
    *{1}{>{\centering\arraybackslash}p{0.5cm}}
    *{4}{>{\centering\arraybackslash}p{1.8cm}}
    *{1}{>{\centering\arraybackslash}p{0.7cm}}}
    PJ16~\cite{park2016robust} &
    LZ20~\cite{li2020multi}  &
    Ours & GT & &
    PJ16~\cite{park2016robust} &
    LZ20~\cite{li2020multi}  &
    Ours & GT &
    \\
    \cmidrule(lr){1-4}\cmidrule(lr){6-9}
    \multicolumn{4}{c}{BEAR} & &\multicolumn{4}{c}{READING} 
    \end{tabular}}}
    \vspace{-0.3em}
    \caption{Qualitative results of shape and normal on \mvdiligentdata.} 
    \label{fig:ps_normal}
\end{figure}


\begin{table}[t] \centering
    \captionof{table}{Comparison of novel view rendering on the \mvdiligentdata.}
    \label{tab:nrf_real}
    
\resizebox{\textwidth}{!}{
\begin{threeparttable}
\begin{tabular}{l|*{3}{c}|*{3}{c}|*{3}{c}|*{3}{c}|*{3}{c}}
    \toprule
    \multicolumn{1}{c|}{} & \multicolumn{3}{c}{BEAR} & \multicolumn{3}{c}{BUDDHA} & \multicolumn{3}{c}{COW} & \multicolumn{3}{c}{POT2} & \multicolumn{3}{c}{READING}
    \\
    Method & PSNR$\uparrow$ & SSIM$\uparrow$ & LPIPS$\downarrow$ 
     & PSNR$\uparrow$ & SSIM$\uparrow$ & LPIPS$\downarrow$
      & PSNR$\uparrow$ & SSIM$\uparrow$ & LPIPS$\downarrow$
       & PSNR$\uparrow$ & SSIM$\uparrow$ & LPIPS$\downarrow$
        & PSNR$\uparrow$ & SSIM$\uparrow$ & LPIPS$\downarrow$
    \\
    \hline
NeRF~\cite{mildenhall2020_nerf_eccv20} 
& 30.98  & 0.9887  & 1.09 
& 29.31  & 0.9664  & 2.80 
& 33.03  & 0.9907  & 0.54 
& 32.52  & 0.9842  & 1.41 
& 26.58  & 0.9664  & 1.84 
\\
KB22~\cite{kaya2022neural}
& 39.63  & \textbf{0.9960}  & \textbf{0.24 }
& 33.62  & \textbf{0.9844}  & 0.56 
& 31.38  & 0.9800  & 0.87 
& 33.39  & 0.9767  & 1.02 
& 22.45  & 0.9560  & 2.11 
\\
UNISURF~\cite{oechsle2021unisurf} 
& 40.13  & 0.9954  & 0.40 
& 30.98  & 0.9707  & 1.98 
& 40.17  & 0.9953  & 0.39 
& 43.06  & 0.9954  & 0.50 
& 22.19  & 0.9579  & 2.29 
\\
NeRFactor~\cite{zhang2021physg}
& 29.28  & 0.9791  & 2.38 
& 26.34  & 0.9385  & 6.33 
& 27.60  & 0.9630  & 1.67 
& 32.32  & 0.9738  & 1.84 
& 25.62  & 0.9468  & 2.99 
\\
NeRD~\cite{boss2021neural} 
& 26.24  & 0.9661  & 3.56 
& 20.94  & 0.8701  & 6.25 
& 23.98  & 0.8914  & 1.97 
& 26.34  & 0.8422  & 2.19 
& 20.13  & 0.9023  & 3.90 
\\
PhySG~\cite{zhang2021physg} 
& 34.01  & 0.9841  & 1.60 
& 29.64  & 0.9594  & 2.65 
& 34.38  & 0.9856  & 1.02 
& 35.92  & 0.9814  & 1.23 
& 24.19  & 0.9531  & 2.88 
\\
\hline
Ours
& \textbf{41.58}  & 0.9959  & 0.31 
& \textbf{33.73}  & 0.9829  & \textbf{0.54 }
& \textbf{42.39}  & \textbf{0.9962}  & \textbf{0.22} 
& \textbf{45.44}  & \textbf{0.9960}  & \textbf{0.15 }
& \textbf{30.47}  & \textbf{0.9808}  & \textbf{0.75 }
\\
    \bottomrule
\end{tabular}
\end{threeparttable}
}
    \vspace{1em}
    \captionof{table}{Shape reconstruction results of neural rendering methods on both real and synthetic datasets. We use both \syndataEnv and \syndataPS for evaluation.}
    \label{tab:nrf_syn_shape}

\resizebox{\textwidth}{!}{
\setlength{\tabcolsep}{2pt}
\begin{tabular}{l*{5}{|*{2}{c}}|*{4}{|*{2}{c}}}
    \toprule
    & \multicolumn{10}{c||}{DiLiGenT-MV} & \multicolumn{8}{c}{Synthetic} \\
    & \multicolumn{2}{c|}{BEAR} & \multicolumn{2}{c|}{BUDDHA} & \multicolumn{2}{c|}{COW} & \multicolumn{2}{c|}{POT2} & \multicolumn{2}{c||}{READING} & \multicolumn{4}{c|}{BUNNY} &  \multicolumn{4}{c}{ARMADILLO}
    \\
     Method  
     &\multirow{2}{*}{MAE$\downarrow$} &\multirow{2}{*}{CD$\downarrow$}
     &\multirow{2}{*}{MAE$\downarrow$} &\multirow{2}{*}{CD$\downarrow$}
     &\multirow{2}{*}{MAE$\downarrow$} &\multirow{2}{*}{CD$\downarrow$}
     &\multirow{2}{*}{MAE$\downarrow$} &\multirow{2}{*}{CD$\downarrow$}
     &\multirow{2}{*}{MAE$\downarrow$} &\multirow{2}{*}{CD$\downarrow$}
     &\multicolumn{2}{c|}{MAE$\downarrow$} &  \multicolumn{2}{c|}{CD$\downarrow$}
     &\multicolumn{2}{c|}{MAE$\downarrow$} &  \multicolumn{2}{c}{CD$\downarrow$}
    \\
    & & & & & & & & & & & PS & Env & PS & Env & PS & Env & PS & Env \\
    \hline
NeRF~\cite{mildenhall2020_nerf_eccv20}	& 73.90	& 66.68	& 59.89	& 29.28	& 55.14	& 70.07	& 69.71	& 42.28	& 55.75	& 48.26	& 49.42	& 48.22	& 19.67	& 20.09	& 44.27	& 41.54	& 20.95	& 24.34	\\
KB22~\cite{kaya2022neural}	& 53.19	& 66.18	& 39.72	& 17.92	& 85.11	& 82.43	& 87.30	& 63.82	& 70.13	& 86.79	& 31.64	& 37.00	& 9.61	& 18.32		& 35.88	& 51.08	& 14.50	& 7.58	\\
UNISURF~\cite{oechsle2021unisurf}	& 6.48	& 9.24	& 17.11	& 9.83	& 8.25	& 13.25	& 13.05	& 10.21	& 19.72	& 62.89	& 10.00	& 11.46	& 6.89	& 8.74	& 8.12	& 10.12	& 3.76	& 3.96	\\
NeRFactor~\cite{zhang2021physg}	& 12.68	& 26.21	& 25.71	& 26.97	& 17.87	& 50.65	& 15.46	& 29.00	& 21.24	& 47.36	& 21.97	& 21.50	& 17.29	& 18.29		& 36.07	& 19.27	& 7.86	& 18.44	\\
NeRD~\cite{boss2021neural}	& 19.49	& 13.90	& 30.41	& 18.54	& 33.18	& 38.62	& 28.16	& 9.00	& 30.83	& 30.05	& 17.51	& 19.12	& 11.08	& 13.44		& 19.36	& 18.43	& 7.02	& 9.34	\\
PhySG~\cite{zhang2021physg}	& 11.22	& 19.07	& 26.31	& 21.66	& 11.53	& 22.23	& 13.74	& 32.29	& 25.74	& 46.90	& 23.53	& 25.66	& 22.23	& 21.87		& 14.46	& 19.53	& 8.27	& 12.08	\\
Ours	& \textbf{3.21}	& \textbf{7.24}	& \textbf{10.10}	& \textbf{8.93}	& \textbf{4.08}	& \textbf{11.33}	& \textbf{5.67}	& \textbf{5.76}	& \textbf{8.83}	& \textbf{12.83}	& \textbf{5.14}	& --	& \textbf{5.32}	& --	& \textbf{5.18}	& --	& \textbf{3.61}	& --	\\
\bottomrule
\end{tabular}
}
    \vspace{1em}
    \\
    \includegraphics[width=0.95\textwidth]{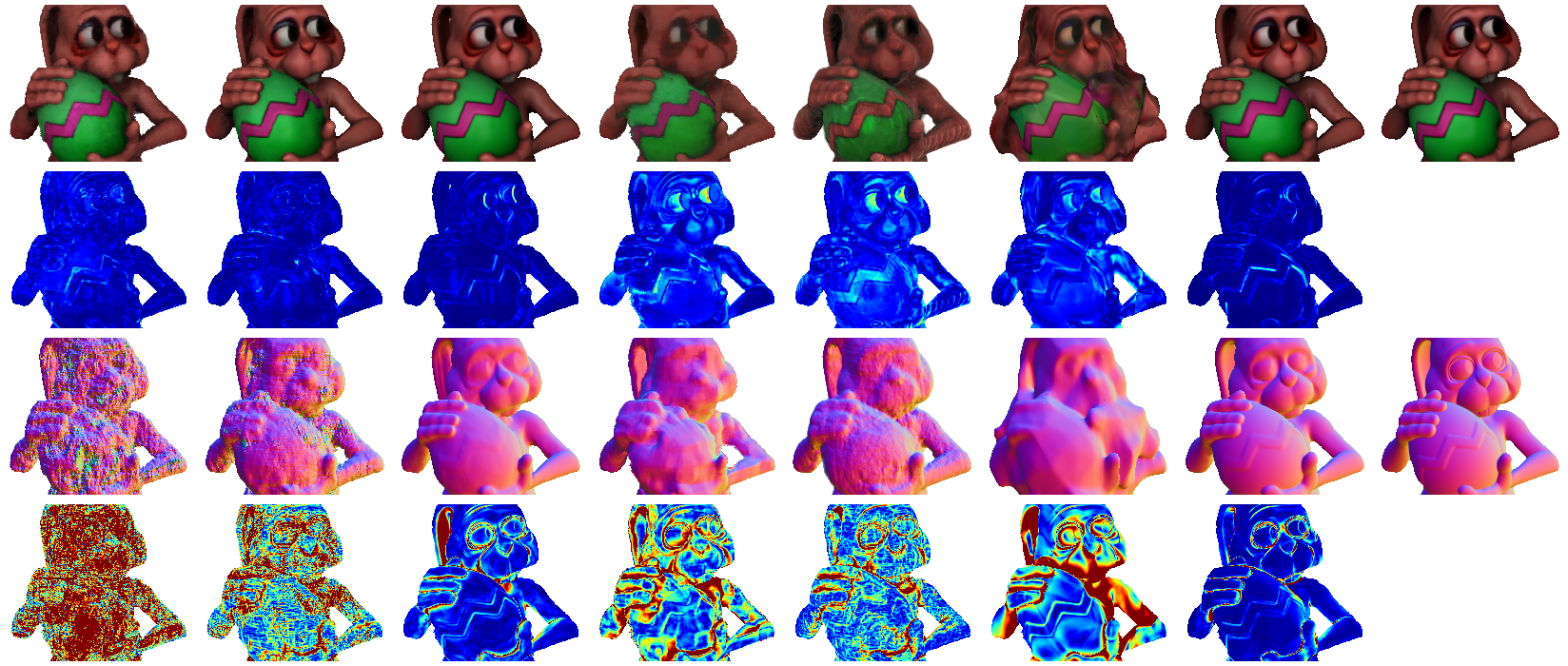}
    \raisebox{0.96\height}{
        \makebox[0.02\textwidth]{
            \makecell{
                \includegraphics[width=0.02\textwidth]{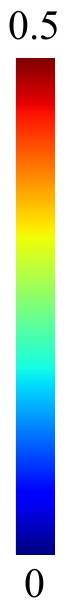}\\[10pt]
                \includegraphics[width=0.02\textwidth]{images/colorbar.pdf}
            }
        }
    }\\
    \vspace{-1em}
    \subfloat{\resizebox{\textwidth}{!}{
    \begin{tabular}{*{8}{>{\centering\arraybackslash}p{2cm}}
    *{1}{>{\centering\arraybackslash}p{1cm}}}
    NeRF~\cite{mildenhall2020_nerf_eccv20} &
    KB22~\cite{kaya2022neural}&
    UNISURF~\cite{oechsle2021unisurf}&
    NeRFactor~\cite{zhang2021physg}&
    NeRD~\cite{boss2021neural} &
    PhySG~\cite{zhang2021physg} &
    Ours & GT &
    \end{tabular}}}\\[3pt]
    \vspace{-0.5em}
    \captionof{figure}{Novel view rendering and normal estimation results on the synthetic dataset.} \label{fig:nerf_compare_syn}
\end{table}

\Tref{tab:ps_shape} summarizes the quantitative comparison. The results of Chamfer distance are calculated with ground truth shape. The normal MAE results of DiLiGenT test views are calculated inside the intersection of input mask and our predicted mask. We can see that our method outperforms previous methods in 4 out of 5 objects by a large margin in terms of both two metrics. In particular, our method significantly reduces the Chamfer distance of the most challenging \emph{READING} object from 22.62mm to 12.35mm. On average, our method improves Chamfer distance by 33$\%$, and normal MAE by 6$\%$. Such remarkable performance proves that our method is much better in shape reconstruction than existing approaches. We notice our results of the \emph{COW} object are slightly worse.  Detailed discussions are in the supplementary material.

To show the advantages of our method intuitively, we present two visual example comparisons in \Fref{fig:ps_normal}. We show the easiest \emph{BEAR} with a smooth surface and the most challenging \emph{READING} with many wrinkles. It can be observed that our results have fewer noises than LZ20 and PJ16 on the smooth surface of \emph{BEAR}, and our reconstruction of the detailed wrinkles of \emph{READING} is much better. It demonstrates that our method is superior in reconstructing both smooth and rugged surfaces.

\subsection{Comparison with Neural Rendering Based Methods}
We also compared our method with existing neural rendering methods, including
NeRF~\cite{mildenhall2020_nerf_eccv20}, KB22~\cite{kaya2022neural}, UNISURF~\cite{oechsle2021unisurf}, PhySG~\cite{zhang2021physg}, NeRFactor~\cite{zhang2021physg}, and NeRD~\cite{boss2021neural}.
Among them, the first three methods can only support novel-view rendering and cannot perform scene decomposition.

For KB22~\cite{kaya2022neural}, we re-implemented it by adding the normals estimated by PS method as input to NeRF following their paper.
For other methods, we used their released codes for experiments.

\begin{table}[t] \centering
    \captionof{table}{Analysis on normal supervision.}
    \label{tab:analysis_normal}
    
\resizebox{\textwidth}{!}{
\begin{tabular}{l|*{6}{c}|*{6}{c}}
    \toprule
    \multicolumn{1}{c|}{} & \multicolumn{6}{c|}{Normal MAE$\downarrow$}& \multicolumn{6}{c}{Chamfer Dist$\downarrow$ }
    \\
    Method &  BEAR & BUDDHA & COW & POT2 & READING & BUNNY
     &  BEAR & BUDDHA & COW & POT2 & READING & BUNNY
    \\
    \hline
NeRF
& 73.45  & 59.62  & 55.10  & 69.41  & 55.55  & 49.24  & 66.68  & 29.28  & 70.07  & 42.28  & 48.26  & 19.67  \\
NeRF$^{+N}$
& 7.03  & 13.50  & 8.26  & 7.93  & 14.01  & 12.04  & 16.02  & 9.65  & 12.04  & 7.25  & 13.31  & 5.44  \\
\hline
UNISURF
& 6.51  & 17.13  & 8.26  & 13.04  & 19.68  & 10.04  & 9.24  & 9.83  & 13.25  & 10.21  & 62.89  & 6.89  \\
UNISURF$^{+N}$
& \textbf{4.26}  & \textbf{11.29}  & \textbf{5.05}  & \textbf{6.37}  & \textbf{9.58}  & \textbf{7.76}  & \textbf{7.24}  & \textbf{
8.93}  & \textbf{11.33}  & \textbf{5.76}  & \textbf{12.83}  & \textbf{5.32}  \\
    \bottomrule
\end{tabular}
}
    \\
\vspace{1em}
    \includegraphics[width=0.8\textwidth]{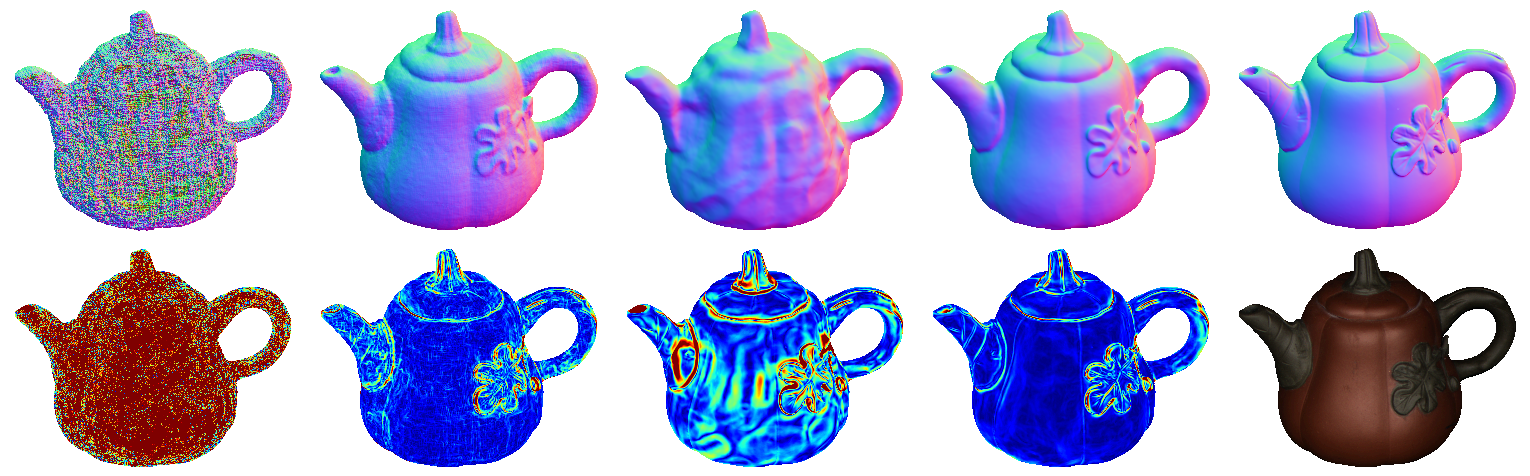}
    \includegraphics[width=0.025\textwidth]{images/colorbar.pdf} \\
    \vspace{-1em}
    \subfloat{\resizebox{0.825\textwidth}{!}{
    \begin{tabular}{*{5}{>{\centering\arraybackslash}p{2cm}}
    *{1}{>{\centering\arraybackslash}p{0.4cm}}}
    NeRF & NeRF$^{+N}$ & UNISURF & UNISURF$^{+N}$ & GT &
    \end{tabular}}}
    \vspace{-1em}
    \captionof{figure}{Qualitative results of  estimated surface normal on w/ or w/o normal supervision.} \label{fig:analysis_normal}
\end{table}

\paragraph{Evaluation on \MVDiligentData}
Following the training of our radiance field stage, we use the multi-light averaged image for each view (equivalent to image lit by frontal hemisphere lights) for training and testing the baselines. 
For our method, we rendered directional light images and computed the light-averaged image for each view in testing phases. 
As the light intensity can only be estimated up to an unknown scale, we reported the scale-invariant PSNR, SSIM and LPIPS for all methods by first finding a scalar to rescale the rendered image using least-square.

\Tref{tab:nrf_real} and \Tref{tab:nrf_syn_shape} show that our method achieves the best rendering and normal estimation results, and significantly outperforms the existing best method UNISURF~\cite{oechsle2021unisurf} for shape reconstruction. 
We attribute the success in view rendering to the faithful decomposition of shape, material, and light components and our shadow-aware design. The multi-light images provide abundant high-frequency information related to the shape and material, which ensures high rendered quality under diverse illuminations.
 
\paragraph{Evaluation on Synthetic Data}
Since existing neural rendering methods mainly assume a fixed environment lighting, we additionally evaluate on the synthetic dataset that is rendered with both PS lights (\syndataPS) and environment light (\syndataEnv) to investigate the shape reconstruction performance under different light conditions.

The results in \Tref{tab:nrf_syn_shape} reports similar geometry reconstruction performance for all baselines methods in both PS and environment lighting.
They failed to reconstruct accurate surface in both light conditions.
This may be due to the sparse input views, and the ambiguity in shape and material joint estimation, especially for non-textured regions, which exists in both PS and Env lightings.
In contrast, our method successfully estimates more faithful geometry.
We also show the qualitative comparison in \fref{fig:nerf_compare_syn}.
Our method achieves the best performance on both re-rendered quality and reconstructed surface normal, especially for details such as eye regions.

\begin{table}[t] \centering
\begin{minipage}[t]{0.48\linewidth}
        \caption{Normal improvement. }
        \label{tab:improve_normal}

\resizebox{\textwidth}{!}{
\begin{tabular}{l|*{6}{c}}
    \toprule
    Method & BEAR & BUDDHA & COW & POT2 & READING & BUNNY
    \\
    \hline
SDPS-Net & 7.52  & 11.47  & 9.57  & 7.98  & 15.94  & 10.65 \\
Stage I & 4.26  & 11.29  & 5.05  & 6.39  & 9.58  & 7.79 \\
Ours & \textbf{3.25}  & \textbf{10.20}  & \textbf{4.12}  & \textbf{5.73}  & \textbf{8.87}  & \textbf{5.24}  \\
\bottomrule
\end{tabular}
}
\end{minipage}\hfill
\begin{minipage}[t]{0.48\linewidth}
        \caption{Light improvement. }
        \label{tab:improve_light}
        
\resizebox{\textwidth}{!}{
\begin{tabular}{l|*{6}{c}}
    \toprule
    Method & BEAR & BUDDHA & COW & POT2 & READING & BUNNY
    \\
    \hline
SDPS-Net & 4.90 & 7.17 & 8.55 & 4.73 & 9.09 & 8.96 \\
Ours & \textbf{2.27} &\textbf{ 2.75} & \textbf{2.59} & \textbf{2.89} & \textbf{4.26} & \textbf{1.53} \\
    \bottomrule
\end{tabular}
}
\end{minipage}
\vspace{0.5em}
    \caption{Ablation study on Stage II.}
    \label{tab:ablation}

\resizebox{\textwidth}{!}{
\begin{tabular}{l|*{3}{c}|*{3}{c}|*{3}{c}|*{3}{c}|*{3}{c}|*{3}{c}}
    \toprule
    \multicolumn{1}{c|}{} 
    & \multicolumn{3}{c|}{BEAR} & \multicolumn{3}{c|}{BUDDHA} & \multicolumn{3}{c|}{COW} & \multicolumn{3}{c|}{POT2} & \multicolumn{3}{c|}{READING} & \multicolumn{3}{c}{BUNNY} 
    \\
    Method & PSNR$\uparrow$   & SSIM$\uparrow$ & MAE$\downarrow$ 
    & PSNR$\uparrow$   & SSIM$\uparrow$ & MAE$\downarrow$ 
    & PSNR$\uparrow$   & SSIM$\uparrow$ & MAE$\downarrow$ 
    & PSNR$\uparrow$   & SSIM$\uparrow$ & MAE$\downarrow$ 
    & PSNR$\uparrow$   & SSIM$\uparrow$ & MAE$\downarrow$ 
    & PSNR$\uparrow$   & SSIM$\uparrow$ & MAE$\downarrow$ 
    \\
    \hline
fixed-light
& 34.51  & 0.9812  & 3.62  & 29.48  & 0.9661  & 10.51  & 35.58  & 0.9872  & 4.45  & 39.47  & 0.9845  & 5.84  & 24.58  & 0.9708  & 9.18  & 24.04  & 0.9833  & 6.94  \\
w/o vis
& 33.44  & 0.9794  & 3.57  & 28.06  & 0.9571  & 10.29  & 36.94  & 0.9890  & 4.09  & 39.38  & 0.9837  & 5.75  & 24.49  & 0.9699  & 8.91  & 21.29  & 0.9761  & 5.34  \\
Ours
& \textbf{35.68}  & \textbf{0.9837}  & \textbf{3.25}  & \textbf{29.58}  & \textbf{0.9670}  & \textbf{10.20}  & \textbf{37.06}  & \textbf{0.9890}  & \textbf{4.12}  & \textbf{40.01}  & \textbf{0.9860}  & \textbf{5.73 } & \textbf{24.89}  & \textbf{0.9725}  & \textbf{8.87}  & \textbf{25.88}  & \textbf{0.9871}  & \textbf{5.24}  \\
    \bottomrule
\end{tabular}
}
\end{table}

\begin{figure}[t] \centering
    \includegraphics[width=0.95\textwidth]{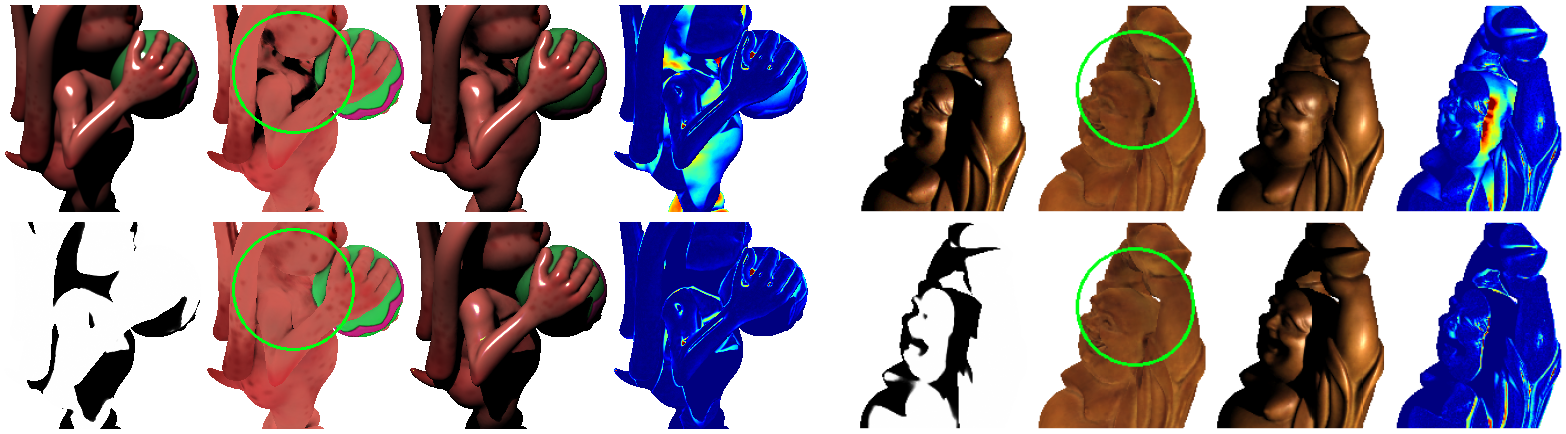}
    \includegraphics[width=0.025\textwidth]{images/colorbar_rgb.pdf} \\
    \vspace{-1em}
    \subfloat{\resizebox{0.95\textwidth}{!}{
    \begin{tabular}{*{4}{>{\centering\arraybackslash}p{2cm}}
    *{1}{>{\centering\arraybackslash}p{0.5cm}}
    *{4}{>{\centering\arraybackslash}p{1.8cm}}
    *{1}{>{\centering\arraybackslash}p{0.5cm}}
    }
    GT/Vis & Albedo & Render & Error & & GT/Vis & Albedo & Render & Error &
    \end{tabular}}}
    \caption{Effectiveness of visibility modeling. (left: BUNNY, right: BUDDHA)} \label{fig:ablation_vis}
\end{figure}

\subsection{Method Analysis}

\paragraph{Effectiveness of Normal Regularization}
Our method exploits multi-light images to infer surface normals to regularize the surface geometry in radiance field. 
It eliminates the ambiguity in density estimation especially for concave-shaped objects.
We show the reconstruction results of before and after adding the normal constraint for both our method (\ie, UNISURF~\cite{oechsle2021unisurf} as our backbone of Stage I) and NeRF~\cite{mildenhall2020_nerf_eccv20} on all objects in \mvdiligentdata and  \emph{BUNNY} in \syndataPS.
As shown in \Tref{tab:analysis_normal} and \fref{fig:analysis_normal}, the introduced normal regularization greatly improves the shape accuracy and recovered surface details for all objects. For example, on the \emph{READING} object, NeRF decreases the normal MAE from $55.55$ to $14.01$ and UNISURF from $19.68$ to $9.58$, which verifies the effectiveness of the normal regularization.

\begin{table}[t]
\begin{minipage}[c]{0.3\textwidth} \centering
    \caption{Quantitative results of different methods when trained with different number of views.} \label{tab:analysis_nviews}
\end{minipage}\hfill
\begin{minipage}[c]{0.68\textwidth}\centering

\resizebox{\textwidth}{!}{
\begin{tabular}{l*{12}{c}}
    \toprule
     \multirow{3}{*}{Method} &\multicolumn{6}{c}{BUNNY} &\multicolumn{6}{c}{ARMADILLO}
    \\
     & \multicolumn{3}{c}{Normal MAE $\downarrow$} & \multicolumn{3}{c}{Chamfer Dist $\downarrow$}& \multicolumn{3}{c}{Normal MAE $\downarrow$} & \multicolumn{3}{c}{Chamfer Dist $\downarrow$}
    \\
    \cmidrule(lr){2-4}\cmidrule(lr){5-7}\cmidrule(lr){8-10}\cmidrule(lr){11-13}
    & 5 & 15 & 30 & 5 & 15 & 30 & 5 & 15 & 30 & 5 & 15 & 30 \\
    \hline
PhySG~\cite{zhang2021physg}
& 34.53  & 25.69  & 27.25  & 36.75  & 21.87  & 42.25  & 30.81  & 19.53  & 20.19  & 49.38  & 12.08  & 16.60  \\
UNISURF~\cite{oechsle2021unisurf} 
& 20.31  & 11.46  & 9.11  & 18.09  & 8.74  & 6.85  & 14.43  & 10.11  & 8.63  & 11.80  & 3.96  & 3.65  \\
Ours
& \textbf{6.96 } & \textbf{5.17}  & \textbf{5.71}  & \textbf{7.71}  & \textbf{5.32}  & \textbf{4.07}  &\textbf{ 6.51 } & \textbf{5.15 } & \textbf{5.02 } & \textbf{5.96}  & \textbf{3.61}  & \textbf{3.31}  \\
    \bottomrule
\end{tabular}
}

\end{minipage}
\end{table}
\begin{figure}[t] \centering
    \vspace{-1em}
    \includegraphics[width=0.96\textwidth]{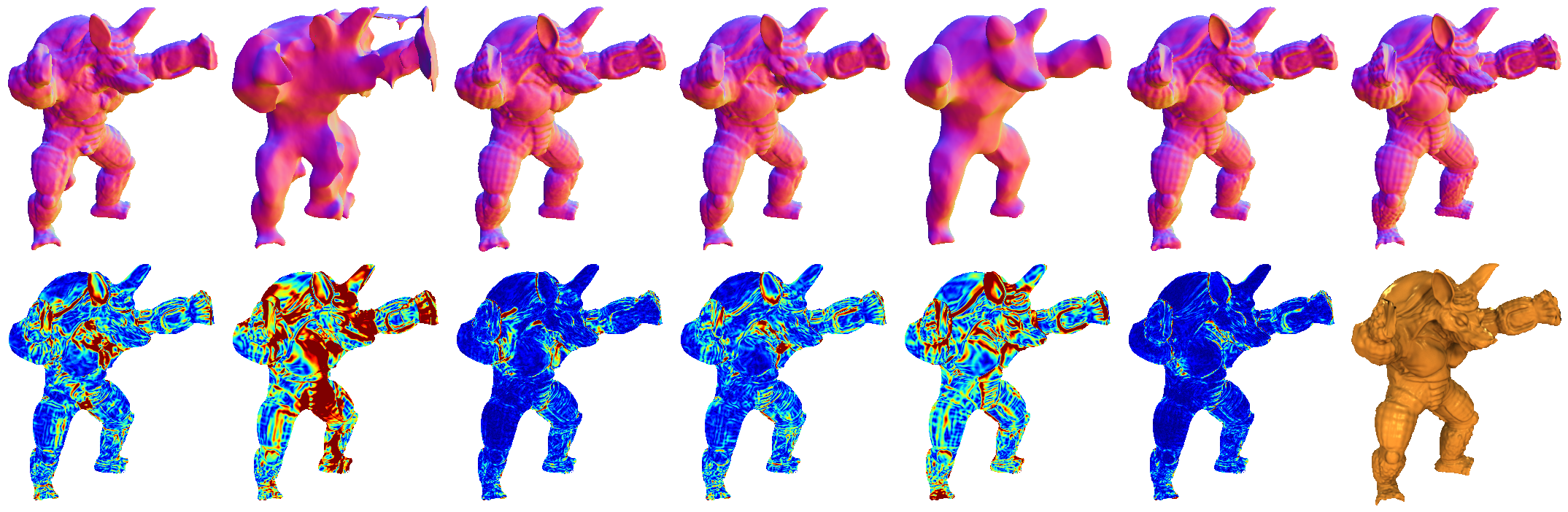}
    \includegraphics[width=0.03\textwidth]{images/colorbar.pdf} \\
    \vspace{-1em}
    \subfloat{\resizebox{\textwidth}{!}{
    \begin{tabular}{*{7}{>{\centering\arraybackslash}p{2cm}}*{1}{p{1cm}}}
    UNISURF~\cite{oechsle2021unisurf}&
    PhySG~\cite{zhang2021physg} &
    Ours & 
    UNISURF~\cite{oechsle2021unisurf}&
    PhySG~\cite{zhang2021physg} &
    Ours & 
    GT &
    \\
    \cmidrule(lr){1-3}\cmidrule(lr){4-6}
    \multicolumn{3}{c}{5 views} &\multicolumn{3}{c}{15 views}
    & &
    \end{tabular}}} 
    \vspace{-0.5em}\captionof{figure}{Qualitative comparison on training with different number of views.} \label{fig:analysis_nview}
    \vspace{1em}
    \captionof{table}{Quantitative results of our method with different number of light directions.}
    \label{tab:analysis_nlight}
    \resizebox{0.8\textwidth}{!}{
\begin{tabular}{c|*{3}{c}|*{3}{c}|*{1}{c}|*{2}{c}}
    \toprule
    \multicolumn{1}{c|}{} & \multicolumn{3}{c|}{Render} & \multicolumn{3}{c|}{Normal MAE$\downarrow$} & \multicolumn{1}{c|}{Shape} & \multicolumn{2}{c}{Light Dir MAE$\downarrow$}
    \\
    N.light & PSNR$\uparrow$ & SSIM$\uparrow$ & LPIPS$\downarrow$  & SDPS  & Stage I   & Ours  & Chamfer Dist$\downarrow$  & SDPS & Ours
    \\
    \hline
1 & 20.21  & 0.9686  & 2.08  & 48.52  & 17.81  & 14.21  & 11.94 & 47.22 & 3.82 \\
2 & 20.69  & 0.9713  & 1.80  & 34.96  & 16.53  & 13.40  & 12.86 & 18.74 & 2.55  \\
4 & 24.78  & 0.9855  & 0.84  & 14.76  & 9.20  & 6.75  & 7.37 & 10.32 & 1.59  \\
8 & 25.84  & 0.9873  & 0.70  & 11.05  & 8.29  & 5.39  & 5.27 & 10.24 & 1.59  \\
16 & 26.00  & 0.9876  & 0.69  & 10.86  & 8.07  & 5.30  & 5.43 & 9.71 & \textbf{1.31}  \\
32 & \textbf{26.03}  & \textbf{0.9880}  & \textbf{0.67}  & 10.69  & 7.84  & 5.21  & \textbf{4.92} & 9.33 & 2.08  \\
96 & 25.90  & 0.9873  & 0.68  & \textbf{10.65}  & \textbf{7.74}  & \textbf{5.18}  & 5.32 &\textbf{ 8.96} & 1.53  \\
    \bottomrule
\end{tabular}
}
\end{figure}

\paragraph{Effectiveness of the Joint Optimization}
Benefiting from the rich shading information in multi-light images and our shadow-aware renderer, we are able to reconstruct faithful surface through joint optimizing normal, BRDFs and lights.
\Tref{tab:improve_normal} shows that joint optimization consistently improves the surface normal accuracy. Besides, the light directions are also refined through joint optimization as \Tref{tab:improve_light} shows.

We also investigated the design of our stage II by ablating the light direction optimization and visibility modeling. 
For ablation study on our own method, we calculate metrics of re-rendered images quality for each light sources instead of light-averaged image.
\Tref{tab:ablation} indicates that either fixing the initialized light direction or removing the visibility modeling will decrease the rendering quality and shape reconstruction accuracy.
In particular, when light visibility is not being considered, the estimated albedo will be entangled with the cast-shadow, leading to inaccurate material estimation (see~\fref{fig:ablation_vis}).

\paragraph{Effect of Training View Numbers}
Since our method can make full use of multi-light image to resolve the depth ambiguity in plain RGB image, it is able to reconstruct high-quality shape just from sparse views. 
To justify it, we compared our method with UNISURF~\cite{oechsle2021unisurf} and PhySG~\cite{zhang2021physg} for surface reconstruction using $5$, $15$, and $30$ training views. 
Note that the baseline methods were trained on \syndataEnv as they assume a fixed environment lighting.
\Tref{tab:analysis_nviews} and \fref{fig:analysis_nview} show the reconstruction results using different number of views. 
Our method achieves satisfying reconstruction results even when only $5$ views are given, whereas other methods all fail when view numbers are insufficient.

\paragraph{Effect of Light Direction Numbers} \label{sec:analysis_nlight} 
While most previous PS methods requires calibrated lights, our method assumes uncalibrated lights and can handle an arbitrary number of lights.
We also conduct experiments to explore how many light numbers are needed for reconstructing high-quality shapes, where $15$ views are used.
Similar to ablation study, we estimate per-light error for the re-rendered images. For light direction error, we take the mean values of their own used lights.
\Tref{tab:analysis_nlight} shows that increasing the number of lights generally increases the reconstruction accuracy. 
Our full method consistently outperforms SDPS-Net and Stage I under different number of lights.
Given only $4$ images illuminated under directional light for each view, our method achieves results comparable to that uses $96$ images (\eg, $6.75$ vs. $5.18$ for MAE), demonstrating the our method is quite robust to numbers of light directions.

\section{Conclusions}
In this paper, we have introduced a neural inverse rendering method for multi-view photometric stereo under uncalibrated lights.
We first represent an object with neural radiance field whose surface geometry is regularized by the normals estimated from multi-light images.
We then jointly optimize the surface normals, BRDFs, visibility, and lights to minimize the image reconstruction loss based on a shadow-aware rendering layer.
Experiments on both synthetic and real dataset show that our method outperforms existing MVPS methods and neural rendering methods.
Notably, our method is able to recover high-quality surface using as few as $5$ input views.

\paragraph{Limitation}
Although our method has been successfully applied to recover high-quality shape reconstruction for complex real-world objects, it still has the following limitations.
First, we ignore surface inter-reflections in the rendering equation.
Second, we assume a solid object to locate its surface locations, thus cannot handle non-solid objects (\eg, fog).
Last, similar to most of the neural rendering methods, we assume the camera poses are given. 
In the future, we are interested in extending our method to solve the above limitations.

\paragraph{Acknowledgements}
This work was partially supported by the National Key R\&D Program of China (No.~2018YFB1800800), the Basic Research Project No.~HZQB-KCZYZ-2021067 of Hetao Shenzhen-HK S\&T Cooperation Zone, NSFC-62202409, and Hong Kong RGC GRF grant (project\# 17203119).

\clearpage
%
%
\bibliographystyle{splncs04}
\bibliography{ref}

\clearpage

\setcounter{section}{0}
\renewcommand{\thesection}{\Alph{section}}
\section*{Appendix}
\section{More Details for the Proposed Method}

\subsection{More Details for Uncalibrated Photometric Stereo}
We adopted a recent uncalibrated photometric stereo (UPS) method, called SDPS-Net~\cite{chen2019self}, to estimate coarse surface normals and light directions. 
SDPS-Net is trained with a synthetic dataset, and we used the publicly available code and model for inference\footnote{https://github.com/guanyingc/SDPS-Net}.

For each view, SDPS-Net takes the multi-light images and the object mask as input, and estimate a surface normal map, light directions and light intensities.

\subsection{More Details for Stage I}
\vspace{-0.2em}
\paragraph{Network Architecture}
The network architecture of our Stage I is the same as UNISURF~\cite{oechsle2021unisurf}. 

\paragraph{Training Details}
We use Adam as optimizer and set learning rate as $0.0001$. For loss weight, we empirically adopt  $\{1,0.05,0.005\}$ for $\alpha_{1-3}$. 
Different from UNISURF~\cite{oechsle2021unisurf}, we utilized the normals estimated by SDPS-Net~\cite{chen2019self} to regularize the normals derived from the density field.
The normal regularization loss was added after 1K iterations to stabilize the training.
We trained Stage I for 100K iterations, which took about $12$ hours to converge.

\subsection{More Details for Stage II}
\vspace{-0.2em}
\paragraph{Network Architecture} 
We use 4-layer MLPs with width 128 for normal and albedo estimation, and a 2-layer MLPs with width 64 for predicting weights of specular SG basis. An 8-layer MLP with width 256 is used for visibility estimation. We add skip connection for normal, albedo and visibility MLPs at 2-th, 2-th and 4-th layer. We choose ReLU as the activation function.

\begin{figure}[htbp]
    \centering
    \includegraphics[width=\textwidth]{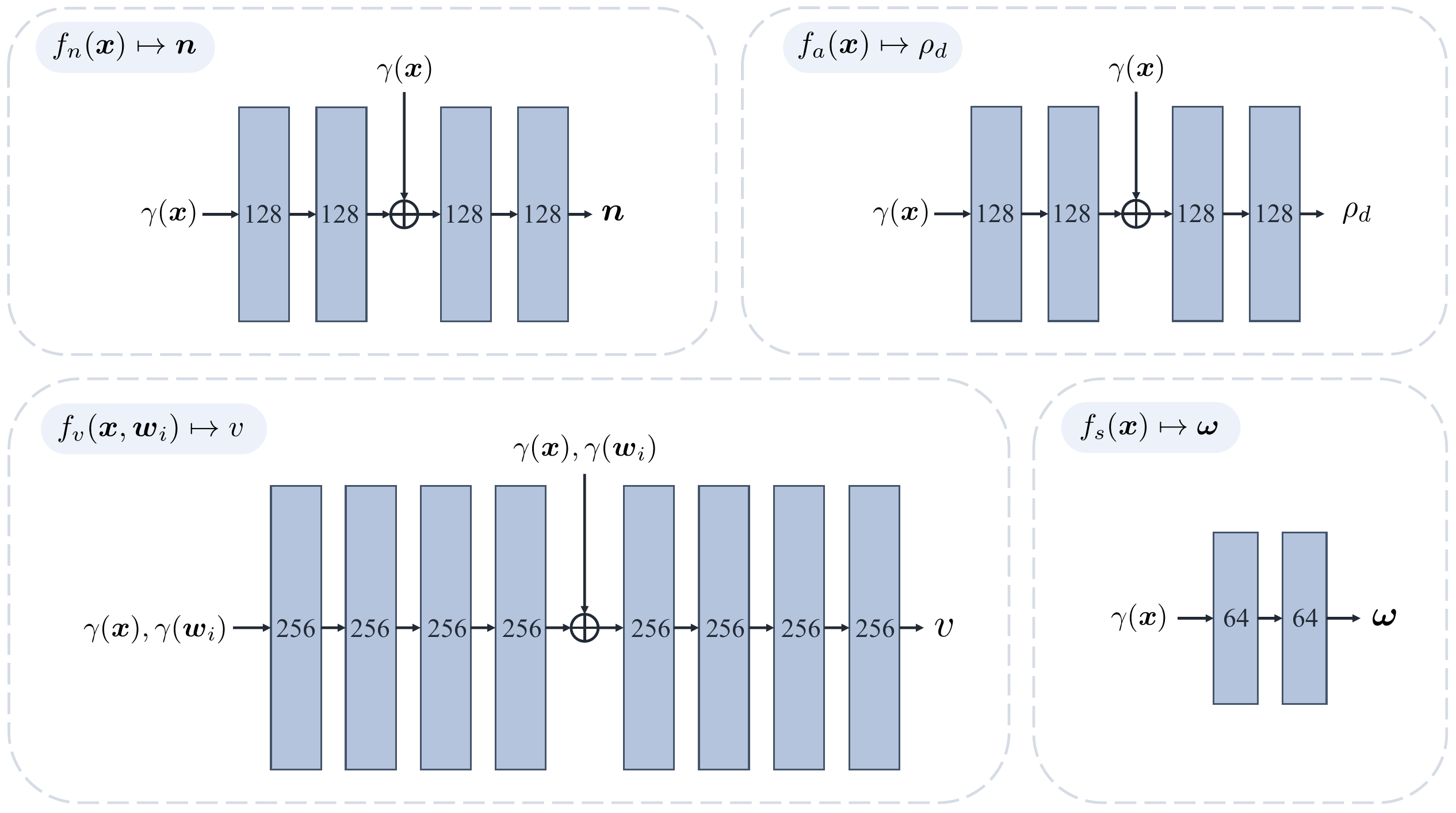}
    \caption{Network architecture of the four MLPs in Stage II.}
    \label{fig:stage2_mlps}
\end{figure}

\Fref{fig:stage2_mlps} shows the detailed network architecture of the four MLPs used in Stage II. We applied positional encoding with 10 frequency components to embed both input point $\boldsymbol{x}$ and light direction $\boldsymbol{w}_i$ into a higher dimensional space. The positional encoding is similar to \cite{mildenhall2020_nerf_eccv20}:
\begin{align}
    \gamma(p)=\left(\sin \left(2^{0} \pi p\right), \cos \left(2^{0} \pi p\right), \cdots, \sin \left(2^{L-1} \pi p\right), \cos \left(2^{L-1} \pi p\right)\right).
\end{align} 
The input of the four MLPs are the encoded point location $\gamma(\boldsymbol{x})$ and encoded light direction $\gamma(\boldsymbol{w}_i)$.

\paragraph{Training Details}
We use Adam as optimizer for stage II and set learning rate as $0.0005$. For loss weight, we empirically adopt  $\{1,1,1,0.05,0.01\}$ for $\beta_{1-5}$.
To stabilize the training process, we first trained the normal MLP and visibility MLP for 5000 iterations, fixing weights of other MLPs and light parameters.
We then jointly trained the normal, visibility, albedo, and specular MLPs, as well as optimize light parameters. 
We trained Stage II for 150K iterations, which took about 10 hours to converge.

\section{More Details for the Comparison}

\subsection{Discussion for the Result on \emph{COW}}
Table 2 of the paper shows that our method performs slightly worse than PJ16~\cite{park2016robust} on \emph{COW} in the metric of Chamfer distances (i.e., 10.21 vs. 9.25).
The main reason is that the camera poses in \mvdiligentdata are located at the upper-hemisphere and slightly look downward, and the objects are placed on a desk with bottom part invisible. 
As a result, accurate reconstruction for the bottom part from images is impossible.
Moreover, the bottom surface of \emph{COW}) is slightly concave, which further enlarges the final mesh error. 

\begin{figure}[htbp]
    \centering
    \includegraphics[width=0.7\textwidth]{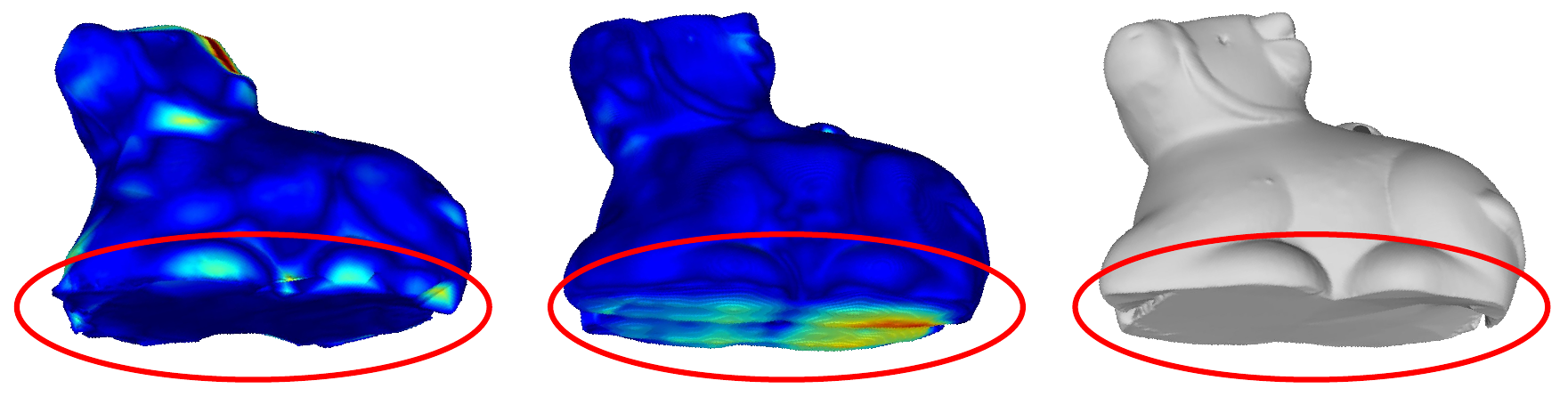}
    \raisebox{0.7\height}{
        \makebox[0.025\textwidth]{
            \makecell{
                \includegraphics[width=0.02\textwidth]{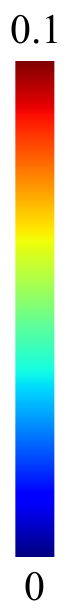}\\[1em]
            }
        }
    }
    \\
    \vspace{-1.0em}
    \makebox[0.23\textwidth]{PJ16~\cite{park2016robust}} 
    \makebox[0.23\textwidth]{Ours} 
    \makebox[0.23\textwidth]{GT} 
    \makebox[0.025\textwidth]{} 
    \\
    \caption{Visualization of mesh reconstruction errors on \emph{COW}. We can see that our method has larger errors on the bottom region, which is invisible in observed images and cannot be reconstructed accurately.}
    \label{fig:cow_bottom}
    \vspace{-1em}
\end{figure}

\Fref{fig:cow_bottom} visualizes the mesh reconstruction error. We can see that our method achieves more accurate reconstruction on the visible surfaces, and most of the error are in the bottom regions.

\subsection{Discussion for the Mask Used in MAE Calculation}
For fair comparison between different methods or analysis cases, we used the overlapped mask region for calculating the metrics. Therefore, there may be some small differences between the values shown in different tables.

\begin{figure*}[htbp] \centering
    \newcommand{\hwidth}{1pt}
    \newcommand{\imgwidth}{0.15\textwidth}
    \newcommand{\patchwidth}{0.070\textwidth}
    \makebox[\imgwidth]{\tiny LZ20/GT Mask}
    \quad
    \makebox[\imgwidth]{\tiny LZ20/Computed Mask}
    \qquad
    \makebox[\imgwidth]{\tiny Ours/GT Mask} 
    \quad
    \makebox[\imgwidth]{\tiny Ours/Computed Mask} 
    \\
    \input{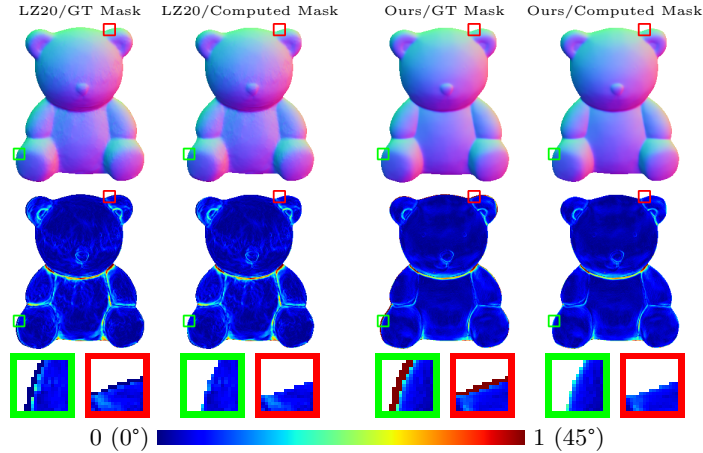}
    \caption{Normal estimation error measured on the ground-truth mask and the computed mask of our method. As the computed mask is generate from the density field, there is a slight misalignment in the boundary region which increases the normal estimation error of our method.} 
    \label{fig:bear_mae}
\end{figure*}

In Table 2 of the paper, we measure the mean angular error (MAE) of the normal estimation using the intersection of input mask and our predicted mask. 
This is because our method reconstructs the full shape in a radiance field and uses projection to get image normals, the boundary regions might not be well aligned with the ground-truth mask. 
\Fref{fig:bear_mae} visualizes that when measures the normal estimation error with the ground-truth mask, our results will have a large error on the boundary region which has a thickness of about $1$ pixel.

We also show the normal MAE results measured on the ground-truth mask in \Tref{tab:ps_normal_gtmask}. Despite having larger errors on the boundary, our method still achieves the lowest average MAE on \mvdiligentdata, which clearly verifies the effectiveness of our method.
\begin{table}[htbp] \centering
    \caption{Results of normal MAE calculated with GT mask.}
    \label{tab:ps_normal_gtmask}

\resizebox{0.6\textwidth}{!}{
\begin{tabular}{l|*{5}{c}|c}
    \toprule
    Method 
    & \emph{BEAR}  & \emph{BUDDHA}   & \emph{COW}   & \emph{POT2}   & \emph{READING}  & \emph{Average} 
    \\
    \hline
PJ16~\cite{park2016robust}
 & 12.63  & 14.58  & 13.24  & 15.31  & 12.23  & 13.60  \Tstrut\\
 LZ20~\cite{li2020multi} 
 & 4.45  & 11.64  & \textbf{4.13}  & 6.79  & \textbf{8.74}  & 7.15  \\
 Ours 
 & \textbf{4.39}  & \textbf{11.31}  & 4.79  & \textbf{6.29}  & 8.89  & \textbf{7.13} \\
    \bottomrule
\end{tabular}
}
    \vspace{-0.5em}
\end{table}

\subsection{More Comparisons with Neural Rendering Methods}
\begin{figure}[htbp]
    \vspace{-1em}
    \centering
    \subfloat{\resizebox{\textwidth}{!}{
    \begin{tabular}{*{8}{>{\centering\arraybackslash}p{2cm}}
    *{1}{>{\centering\arraybackslash}p{1cm}}}
    NeRF~\cite{mildenhall2020_nerf_eccv20} &
    KB22~\cite{kaya2022neural}&
    UNISURF~\cite{oechsle2021unisurf}&
    NeRFactor~\cite{zhang2021physg}&
    NeRD~\cite{boss2021neural} &
    PhySG~\cite{zhang2021physg} &
    Ours & GT &
    \end{tabular}}}\\
    \includegraphics[width=0.95\textwidth]{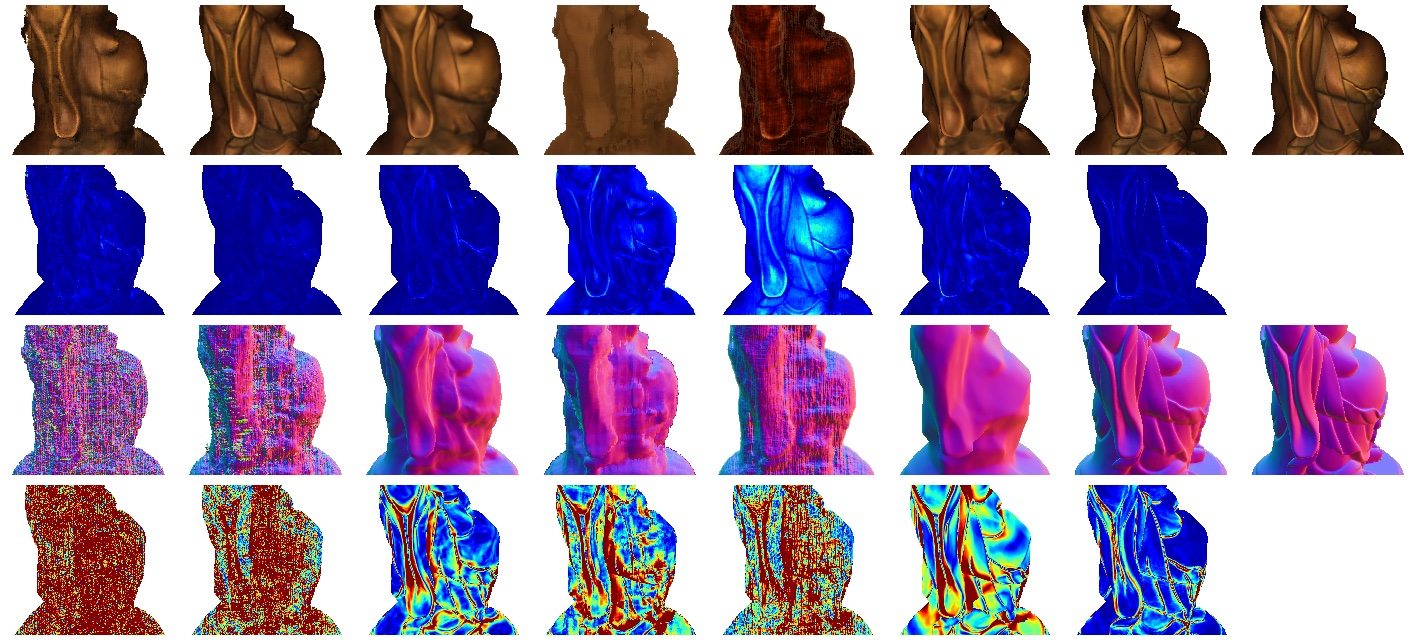}
    \raisebox{0.96\height}{
        \makebox[0.02\textwidth]{
            \makecell{
                \includegraphics[width=0.02\textwidth]{images/colorbar_rgb.pdf}\\[10pt]
                \includegraphics[width=0.02\textwidth]{images/colorbar.pdf}
            }
        }
    }\\
    \caption{More comparison with neural rendering methods on \emph{BUDDHA} from \mvdiligentdata.}
    \label{fig:supp_nrf_compare_buddha}
    \centering
    \subfloat{\resizebox{\textwidth}{!}{
    \begin{tabular}{*{8}{>{\centering\arraybackslash}p{2cm}}
    *{1}{>{\centering\arraybackslash}p{1cm}}}
    NeRF~\cite{mildenhall2020_nerf_eccv20} &
    KB22~\cite{kaya2022neural}&
    UNISURF~\cite{oechsle2021unisurf}&
    NeRFactor~\cite{zhang2021physg}&
    NeRD~\cite{boss2021neural} &
    PhySG~\cite{zhang2021physg} &
    Ours & GT &
    \end{tabular}}}\\
    \includegraphics[width=0.95\textwidth]{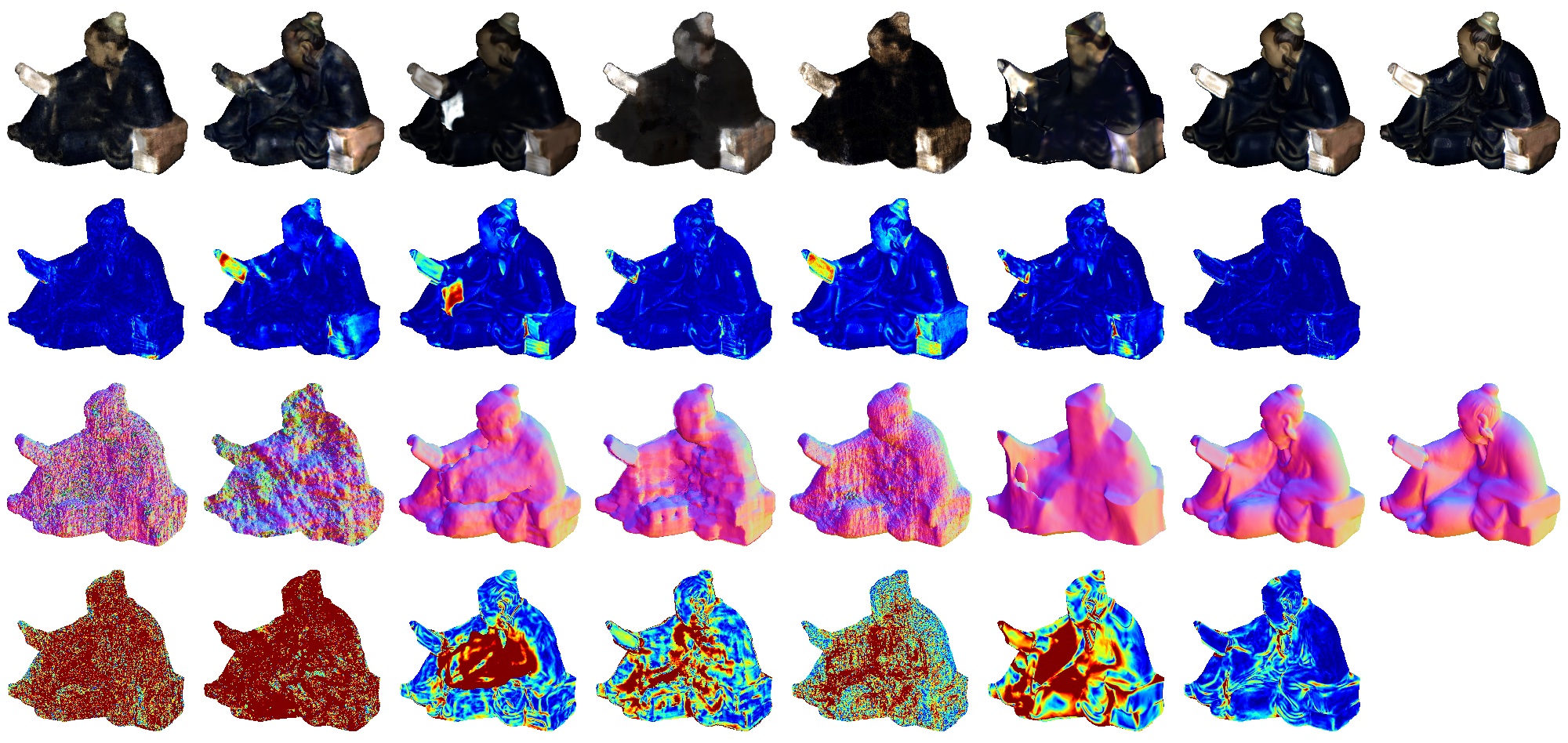}
    \raisebox{0.96\height}{
        \makebox[0.02\textwidth]{
            \makecell{
                \includegraphics[width=0.02\textwidth]{images/colorbar_rgb.pdf}\\[10pt]
                \includegraphics[width=0.02\textwidth]{images/colorbar.pdf}
            }
        }
    }\\
    \caption{More comparison with neural rendering methods on \emph{READING} from \mvdiligentdata.}
    \label{fig:supp_nrf_compare_reading}
    \vspace{-1.5cm}
\end{figure}

\paragraph{More Qualitative Comparisons} \Fref{fig:supp_nrf_compare_buddha} and \Fref{fig:supp_nrf_compare_reading} show the visual comparisons on two objects from \mvdiligentdata. Our method achieves the best rendering and normal reconstruction results.

\clearpage
\section{More Analysis for the Proposed Method}

\subsection{Improvement of Light Estimation}
In Stage II, our method jointly optimizes the lights, normals and BRDFs. \Tref{tab:light_error} shows the refinement of light direction and light intensity over the initialization estimated by SDPS-Net~\cite{chen2019self}.
Our method significantly improves the estimation of SDPS-Net, reducing the the average MAE of light direction from $6.89$ to $2.95$, and average relative error of light intensity from $0.08$ to $0.04$. 
\begin{table}[htbp] \centering
    \caption{Improvement of light direction and intensity estimation compared to SDPS-Net~\cite{chen2019self}.}
    \label{tab:light_error}

\resizebox{\textwidth}{!}{
\begin{tabular}{l|*{5}{c}|c|*{5}{c}|c}
    \toprule
    & \multicolumn{6}{c|}{Light Direction MAE$\downarrow$} 
    & \multicolumn{6}{c}{Light Intensity Error$\downarrow$}\\
    Method & BEAR & BUDDHA & COW & POT2 & READING & Average
    & BEAR & BUDDHA & COW & POT2 & READING & Average
    \\
    \hline
SDPS-Net & 4.90 & 7.17 & 8.55 & 4.73 & 9.09 & 6.89
& 0.10 & 0.06 & 0.11 & \textbf{0.05} & 0.10 & 0.08
\\
Ours & \textbf{2.27} &\textbf{ 2.75} & \textbf{2.59} & \textbf{2.89} & \textbf{4.26}  & \textbf{2.95}
& \textbf{0.04} & \textbf{0.03} & \textbf{0.06} & \textbf{0.05} & \textbf{0.03} & \textbf{0.04}
\\
    
    \bottomrule
\end{tabular}
}

\end{table}

\subsection{Effect of Different Combinations of View and Light Number}
\begin{figure}[htbp] \centering
    \makebox[0.32\textwidth]{5 views} 
    \makebox[0.32\textwidth]{10 views}
    \makebox[0.32\textwidth]{15 views}
    \\
    \includegraphics[width=0.32\textwidth,trim={6cm 11.3cm 6cm 8cm}, clip]{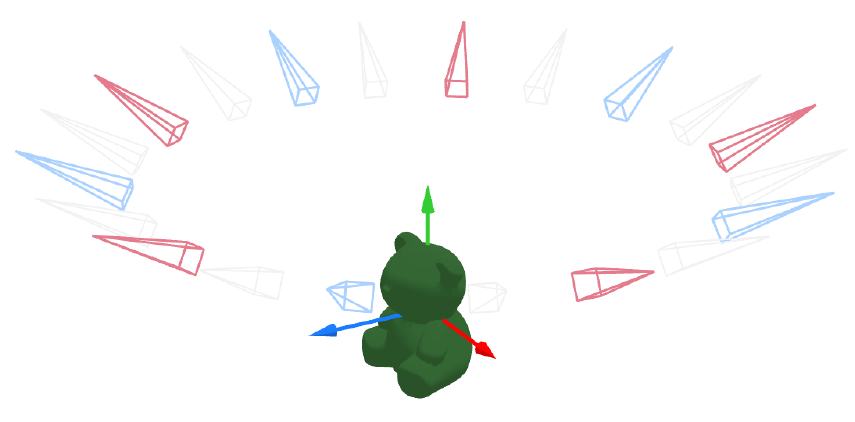}
    \includegraphics[width=0.32\textwidth,trim={6cm 11.3cm 6cm 8cm}, clip]{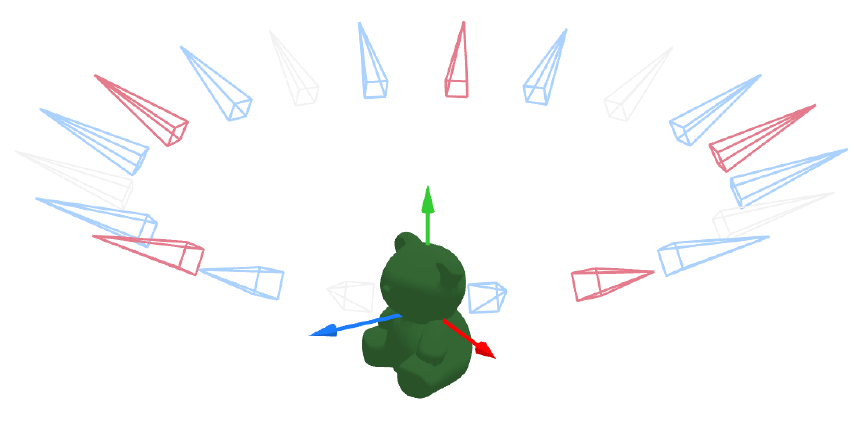}
    \includegraphics[width=0.32\textwidth]{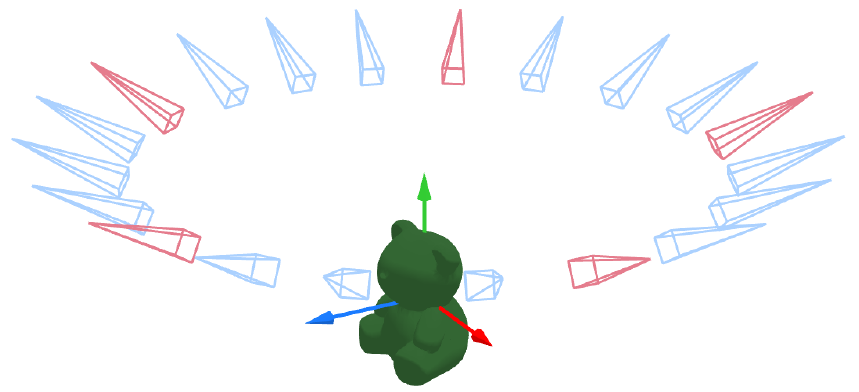}
    \\
    (a) Camera distribution in \mvdiligentdata
    \\[1em]
    \includegraphics[width=0.32\textwidth,trim={5cm 10cm 5cm 5cm}, clip]{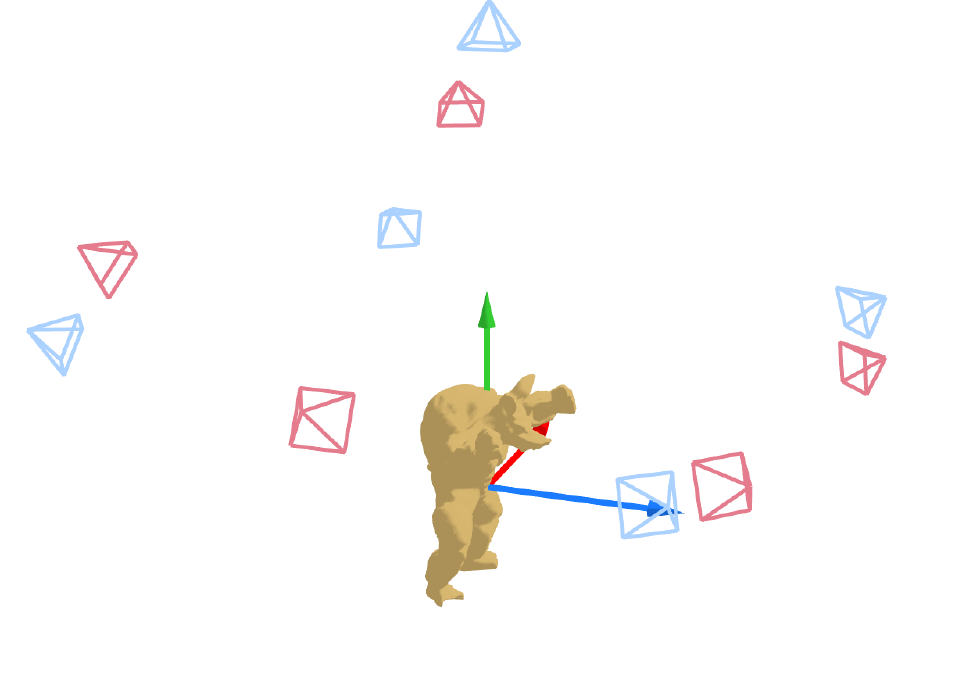}
    \includegraphics[width=0.32\textwidth,trim={5cm 10cm 5cm 5cm}, clip]{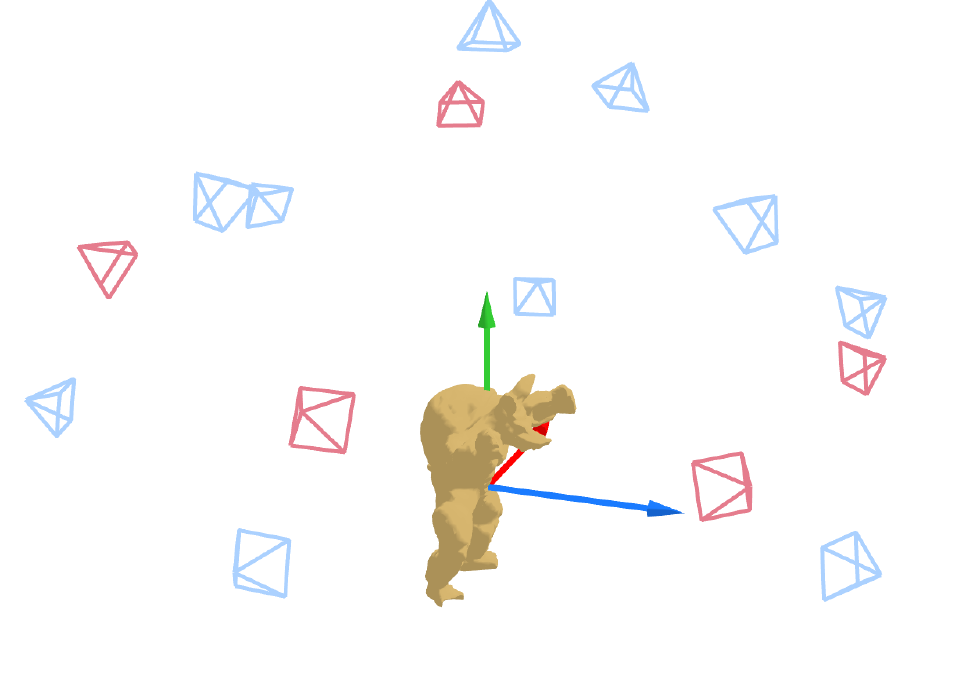}
    \includegraphics[width=0.32\textwidth,trim={5cm 10cm 5cm 5cm}, clip]{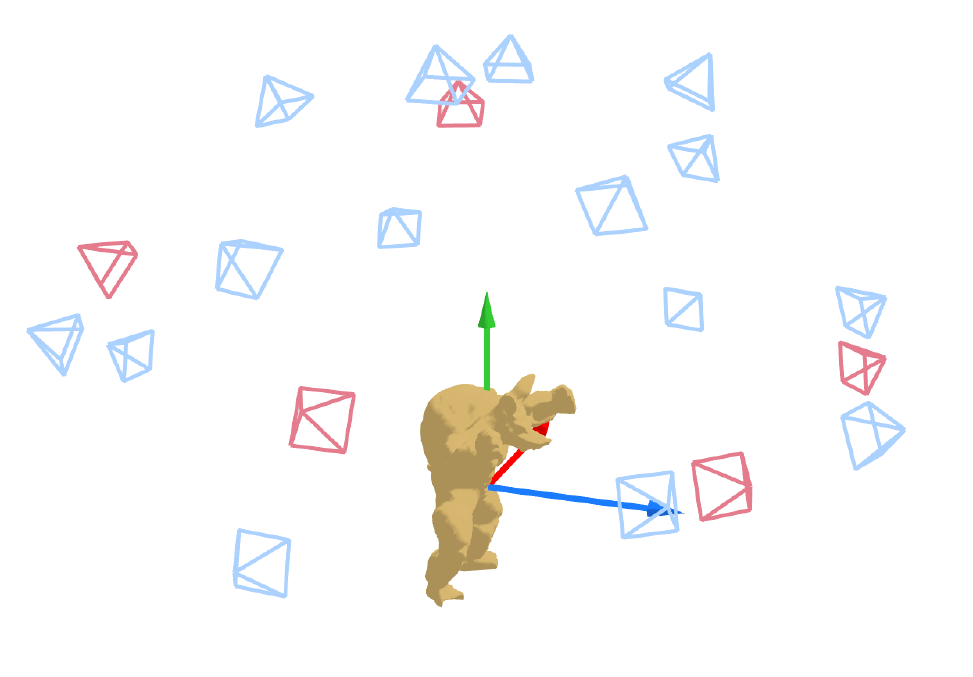}
    \\
    (b) Camera distribution in the synthetic dataset
    \\
    \caption{Visualization of the camera distributions for setups with different input views. Training and testing views are shown in blue and red color, respectively.}
    \label{fig:syn_camera_distribution}
\end{figure}

To further analyze the effect of input view and light numbers, we trained our method with three different view numbers (\ie, $5$, $10$, and $15$). 
The camera distribution of the \mvdiligentdata and the synthetic dataset are shown in \fref{fig:syn_camera_distribution}.

For each input view number, we experimented with four different light numbers (\ie, $2$, $4$, $8$, and $16$). 
Note that the test camera views are the same for all experiments.

\Tref{tab:nlight_nview} shows the normal estimation results of our method on two challenging objects (one real object \emph{READING} and one synthetic object \emph{BUNNY}) with different number of views and lights.
We can see that given more input views and/or light numbers can improve the shape reconstruction results, and our method can achieve robust performance using just a sparse number of views and lights. 

Take results on \emph{READING} as an example, give $5$ views and $8$ lights, our method achieves a MAE of 11.52, which significantly outperforms existing neural rendering methods trained with 15 views (see Table~4 of the paper, where the best performing method UNISURF~\cite{oechsle2021unisurf} achieves a MAE of 19.72).

\begin{table}[htbp] \centering
    \caption{Results of our method on normal estimation error with different combinations of view and light numbers.}
    \label{tab:nlight_nview}

\resizebox{0.6\textwidth}{!}{
\begin{tabular}{c|*{3}{c}|*{3}{c}}
    \toprule
     & \multicolumn{3}{c|}{READING} & \multicolumn{3}{c}{BUNNY}
    \\
    \# Lights & 5 Views & 10 Views & 15 Views  & 5 Views & 10 Views & 15 Views 
    \\
    \hline
2 & 24.48  & 18.33  & 18.52  & 19.20  & 14.52  & 13.39  \Tstrut\\
4 & 14.75  & 11.58  & 11.58  & 9.66  & 7.98  & 6.74  \\
8 & 11.52  & 9.63  & 9.99  & 7.37  & 6.05  & 5.38  \\
16 & 11.09  & 9.17  & 9.44  & 7.05  & 5.20  & 5.28  \\
\bottomrule
\end{tabular}
}
\end{table}

\subsection{Effect of Different BRDF Parameterizations}
As we found it difficult to model the specular effects of real-world objects by directly estimating the roughness parameter of the Microfacet model, we model the specular reflectance with a weighted combination of specular basis following~\cite{hui2017shape,li2022neural}.
\Tref{tab:supp_analysis_microfacet} compares the results of methods using Microfacet model and specular basis on \mvdiligentdata.
We can see that the method using specular basis achieves better results in both image quality and shape reconstruction, which justifies the design of our method.

\begin{table}[htbp] \centering
    \caption{Results of our method on \mvdiligentdata with BRDF parameterizations.}
    \label{tab:supp_analysis_microfacet}
    
\resizebox{\textwidth}{!}{
\setlength{\tabcolsep}{2pt}
\begin{tabular}{l|*{3}{c}|*{3}{c}|*{3}{c}|*{3}{c}|*{3}{c}|*{3}{c}|*{3}{c}}
    \toprule
    BRDF  
    & \multicolumn{3}{c|}{BEAR} & \multicolumn{3}{c|}{BUDDHA} & \multicolumn{3}{c|}{COW} & \multicolumn{3}{c|}{POT2} & \multicolumn{3}{c|}{READING} & \multicolumn{3}{c|}{BUNNY} 
    & \multicolumn{3}{c}{Average} 
    \\
    Modeling & PSNR$\uparrow$   & SSIM$\uparrow$ & MAE$\downarrow$ 
    & PSNR$\uparrow$   & SSIM$\uparrow$ & MAE$\downarrow$ 
    & PSNR$\uparrow$   & SSIM$\uparrow$ & MAE$\downarrow$ 
    & PSNR$\uparrow$   & SSIM$\uparrow$ & MAE$\downarrow$ 
    & PSNR$\uparrow$   & SSIM$\uparrow$ & MAE$\downarrow$ 
    & PSNR$\uparrow$   & SSIM$\uparrow$ & MAE$\downarrow$ 
     & PSNR$\uparrow$   & SSIM$\uparrow$ & MAE$\downarrow$ 
    \\
    \hline
Microfacet
& 35.59  & 0.9836  & \textbf{3.21}  & 28.96  & 0.9641  & 10.28  & 31.19  & 0.9754  & 4.21  & 39.45  & 0.9840  & 5.75  & 24.37  & 0.9692  & 9.07  & 22.76  & 0.9817  & 6.44  & 30.39	& 0.9763	& 6.49 \\
Ours
& \textbf{35.67}  & \textbf{0.9837}  & 3.25  & \textbf{29.58}  & \textbf{0.9670}  & \textbf{10.20}  & \textbf{37.06}  & \textbf{0.9890}  & \textbf{4.12}  & \textbf{40.01}  & \textbf{0.9860}  & \textbf{5.73}  & \textbf{24.89}  & \textbf{0.9725}  & \textbf{8.87}  & \textbf{25.88}  & \textbf{0.9871}  & \textbf{5.24}  & \textbf{32.18}	 & \textbf{0.9809}	& \textbf{6.24} \\
    \bottomrule
\end{tabular}
}

\end{table}

\clearpage
\subsection{Effect of Different Material Types}
\begin{figure}[htbp] \centering
    \subfloat{\resizebox{0.98\textwidth}{!}{
    \begin{tabular}{*{8}{>{\centering\arraybackslash}p{2cm}}
    *{1}{>{\centering\arraybackslash}p{1cm}}}
    GT Image & Our Image & Albedo & Specular & Visibility & Our Normal & Normal Error & GT Normal &
    \\
    \end{tabular}}}
    \\
    \includegraphics[width=0.9\textwidth]{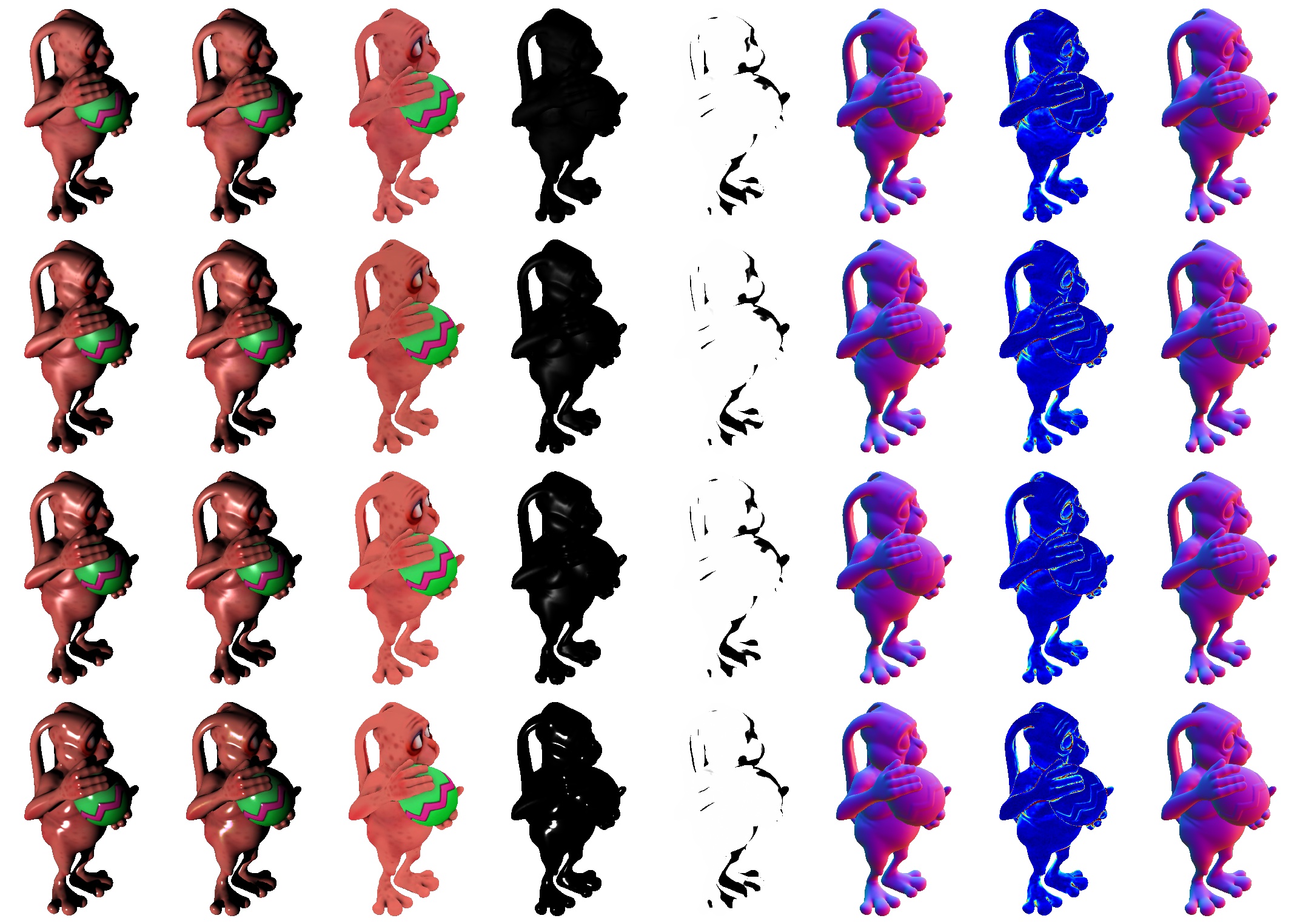}
    \raisebox{0.9\height}{
        \makebox[0.03\textwidth]{
            \makecell{
                \includegraphics[width=0.03\textwidth]{images/colorbar.pdf}\\[9em]
            }
        }
    }
    \caption{Results of our method on materials with different levels of specularity. From top to bottom show the results on object A, B, C, and D. Column 1 shows the ground-truth image, and Column 2--6 show the rendered images, estimated albedo, specular component, visibility, and normals. Column 7--8 shows the normal estimation error and the ground-truth normals.} \label{fig:material_type} 
\end{figure}

To further investigate the results of our method on materials with different levels of specularity, we evaluated our method on \emph{BUNNY} rendered with four strengths of specularity, ranging from diffuse to highly specular (denoted as A, B, C, and D).
\Tref{tab:supp_analysis_material} and \fref{fig:material_type} show the results of our method on these four materials. We can see that our method achieves similar results on these four objects, indicating that our method is robust to different material types.

\begin{table}[htbp] \centering
    \caption{Analysis on the effect of different material types on ``\emphobj{BUNNY}''.}
    \label{tab:supp_analysis_material}

\resizebox{0.8\textwidth}{!}{
\begin{tabular}{c|*{3}{c}|*{3}{c}|*{1}{c}|*{2}{c}}
    \toprule
    \multicolumn{1}{c|}{} & \multicolumn{3}{c|}{Render} & \multicolumn{3}{c|}{Normal MAE$\downarrow$} & \multicolumn{1}{c|}{Shape} & \multicolumn{2}{c}{Light Dir MAE$\downarrow$}
    \\
    Material Type & PSNR$\uparrow$ & SSIM$\uparrow$ & LPIPS$\downarrow$  & SDPS  & Stage I   & Ours  & Chamfer Dist$\downarrow$  & SDPS & Ours
    \\
    \hline
A & 27.79  & 0.9896  & 0.54  & 10.80  & 8.03  & 5.41  & 4.89  & 9.33 & 1.78 \\
B & 27.61  & 0.9896  & 0.53  & 10.31  & 7.58  & 5.38  & 4.77  & 9.30 & 2.57 \\
C & 26.97  & 0.9888  & 0.55  & 10.53  & 7.91  & 5.55  & 5.03 & 9.33 & 2.49  \\
D & 26.02  & 0.9879  & 0.67  & 10.69  & 7.86  & 5.22  & 4.92 & 9.33 & 2.08  \\
\bottomrule
\end{tabular}
}
\end{table}

\clearpage
\section{More Details for the Datasets}

\subsection{Details of the \MVDiligentData}
\begin{figure}[htbp]
    \vspace{-1em}
    \centering
    \includegraphics[width=0.4\textwidth]{images/supp/camera_real_v1.pdf}
    \hspace{1em}
    \includegraphics[width=0.3\textwidth]{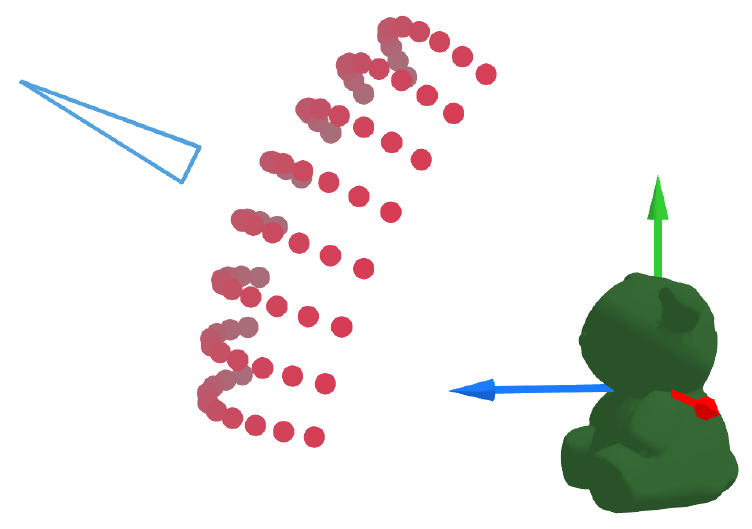}
    \hspace{1em}
    \raisebox{0.9\height}{
        \makebox[0.15\textwidth]{
            \makecell{
                \includegraphics[width=0.15\textwidth]{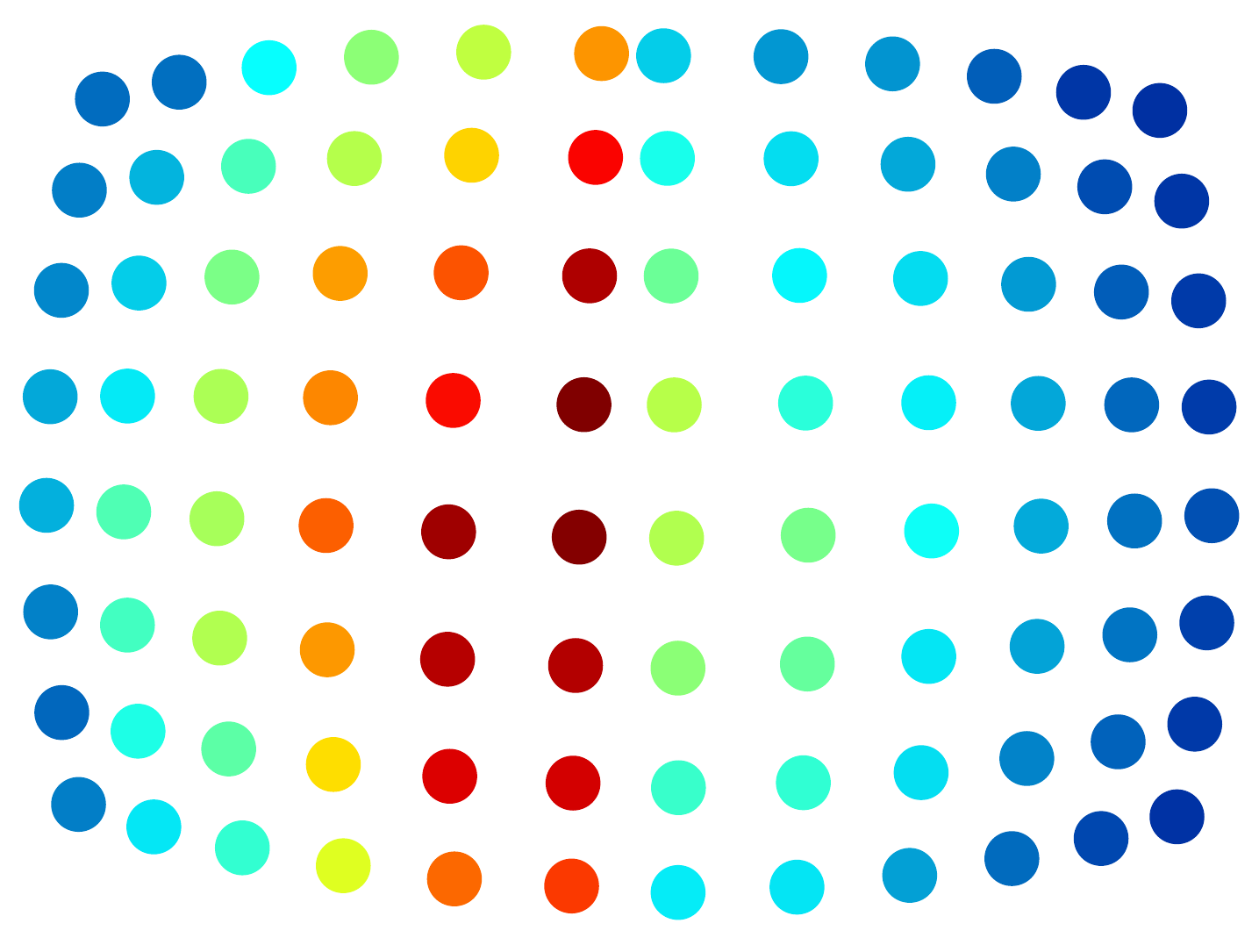}\\[1em]
            }
        }
        \makebox[0.025\textwidth]{
            \makecell{
                \includegraphics[width=0.015\textwidth]{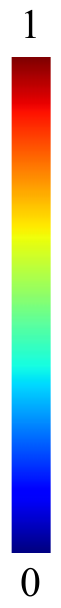}\\[1em]
            }
        }
    }
    \\
    \makebox[0.35\textwidth]{\scriptsize (a) Camera views}
    \makebox[0.6\textwidth]{\scriptsize (b) Distribution of light direction (left) and intensity (right)}\\
    \caption{Camera and light distribution in \mvdiligentdata. The object is scaled up for $5\times$ for easier visualization.
    (a) The object is located at the origin, and the cones indicate the camera poses (blue for training and red for testing views). 
    (b) For each view, $96$ images under different single light directions are captured. At the left, the position of the red point indicates the light direction. 
    At the right, the light intensities are normalized to $[0, 1]$ and visualized with pseudo color.
    }
    \label{fig:camera_distribution_real}
    \vspace{-1.5em}
\end{figure}

\paragraph{Dataset Details}
\mvdiligentdata contains five objects, called \emphobj{BEAR}, \emphobj{BUDDHA}, \emphobj{COW}, \emphobj{POT2}, and \emphobj{READING}. 
For each object, images are captured from $20$ evenly distributed cameras from the same elevation (see \fref{fig:camera_distribution_real}~(a)) . 
For each view, $96$ images are taken under different single directional lights with different light intensities (see \fref{fig:camera_distribution_real}~(b)).

\paragraph{Data Processing}
Following conventional UPS setup, we assume an unknown white directional lights setup.
As we do not have access to ground-truth light intensities, we normalize the original images with light intensities predicted by \cite{chen2019self} when training Stage I. In Stage II, we will refine light intensities during the joint optimization.
To train neural rendering methods on this dataset, we normalized images according to the ground-truth light intensities. 

We cropped the original images with a resolution of $612\times512$ into $400\times400$ images to remove the background for each object.

\definecolor{bisque}{rgb}{1,0.89,0.77}

\subsection{Details of the Synthetic Dataset}
\paragraph{Rendering Details}
We rendered two complex objects (\ie, \emphobj{BUNNY}\footnote{https://www.cgtrader.com/free-3d-print-models/art/sculptures/all-your-egg-are-belong-to-us} and \emphobj{ARMADILLO}\footnote{http://graphics.stanford.edu/data/3Dscanrep/}) under both PS lighting, denoted as \syndataPS, and environment lighting, denoted as \syndataEnv, via Mitsuba \cite{jakob2010mitsuba}. 
The objects were rescaled to within a $[-1,1]$ bounding box, and images with a resolution of $512\times512$ were rendered.

\begin{figure}[h]
    \centering
    \includegraphics[width=0.30\textwidth,trim={5cm 11cm 5cm 5cm}, clip]{images/supp/camera_syn_15.pdf}
    \hspace{2em}
    \includegraphics[width=0.35\textwidth]{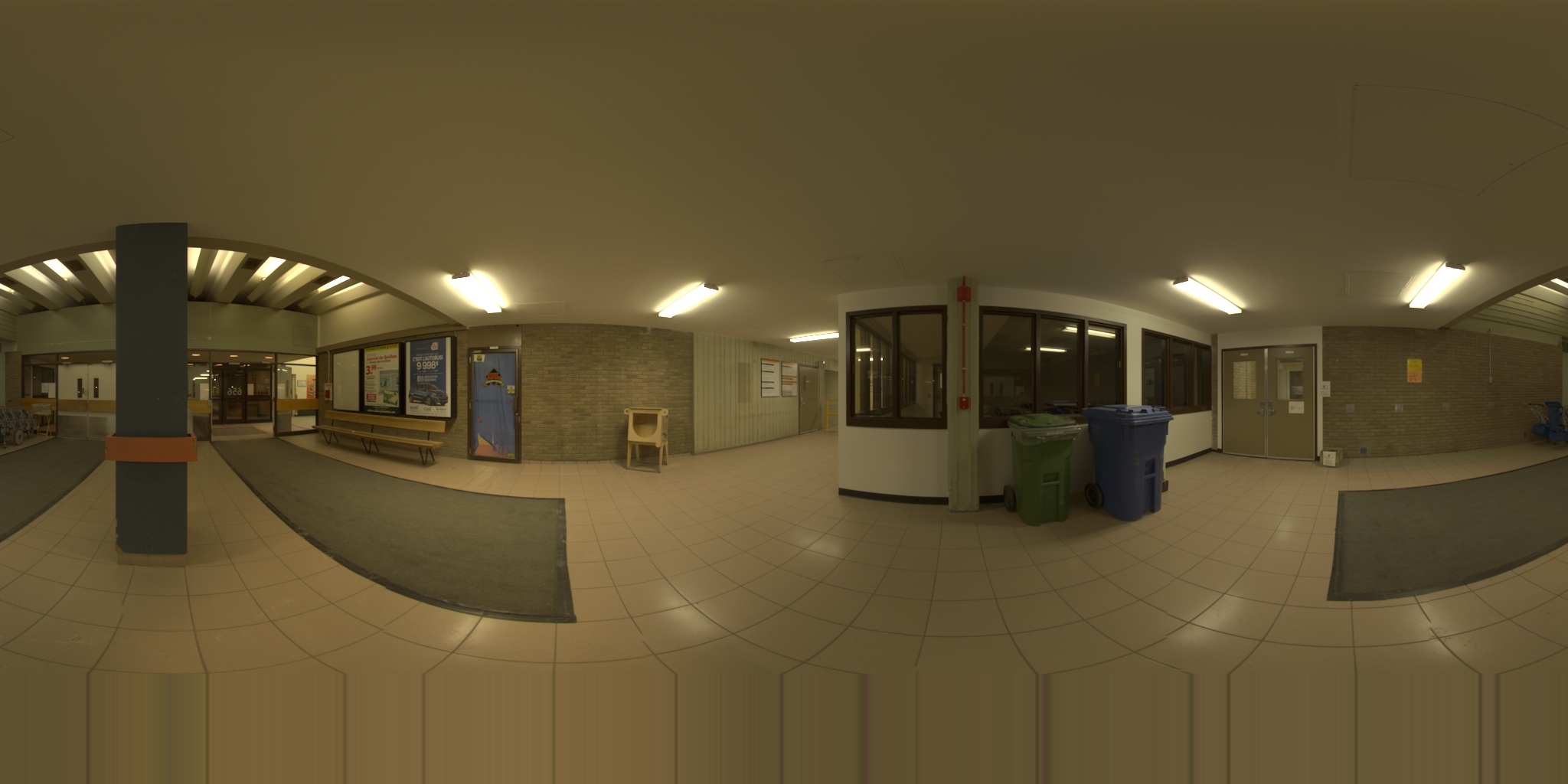}
    \\
    \makebox[0.4\textwidth]{(a) Camera distribution} 
    \makebox[0.55\textwidth]{(b) Environment map used for \syndataEnv}
    \caption{(a) Camera distribution used in the synthetic dataset, where $15$ training views are visualized in blue color and $5$ testing view are in red color. 
    (b) An environment map used for rendering \syndataEnv.}
    \label{fig:syn_15cameras}
    \vspace{-1em}
\end{figure}

\paragraph{Camera \& Light Distribution}
We used the same camera distribution for \syndataEnv and \syndataPS. 
We randomly sampled $20$ camera views on the upper hemi-
sphere, where 15 views for training and 5 views for testing (see~\fref{fig:syn_15cameras}~(a)).
For \syndataPS, we use the same light distribution as \mvdiligentdata for each view, except that we set the same light intensity for each light.
For \syndataEnv, we used an indoor environment map (see~\fref{fig:syn_15cameras}~(b)) for both objects.


\section{Applications}
\begin{figure}[htbp] 
    \vspace{-1.5em}
    \centering
    {\footnotesize
    \subfloat{\resizebox{\textwidth}{!}{
    \begin{tabular}{*{9}{>{\centering\arraybackslash}p{1.8cm}}}
    GT Image & Our Image & Albedo & Specular & Visibility &
     Editing 1 & Editing 2 & Relighting 1 & Relighting 2
     \\
    \end{tabular}}}
    }
    \\
    \includegraphics[width=\textwidth]{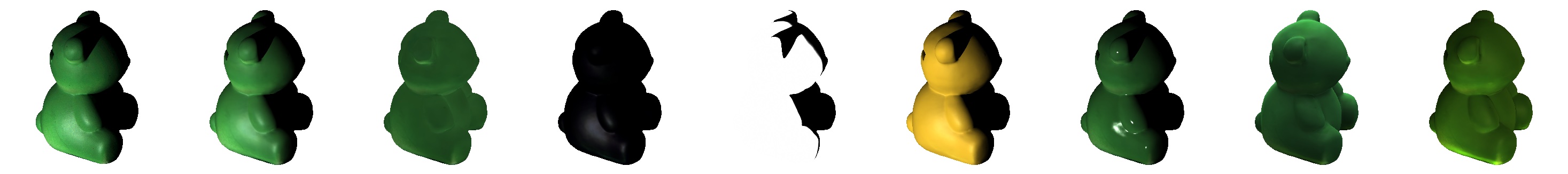}
    \includegraphics[width=\textwidth]{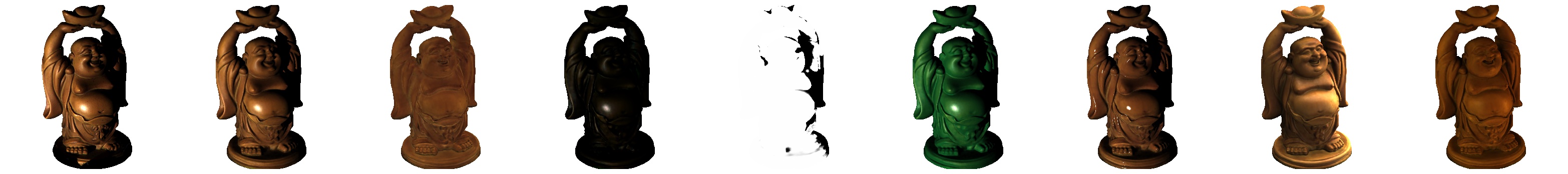}
    \includegraphics[width=\textwidth]{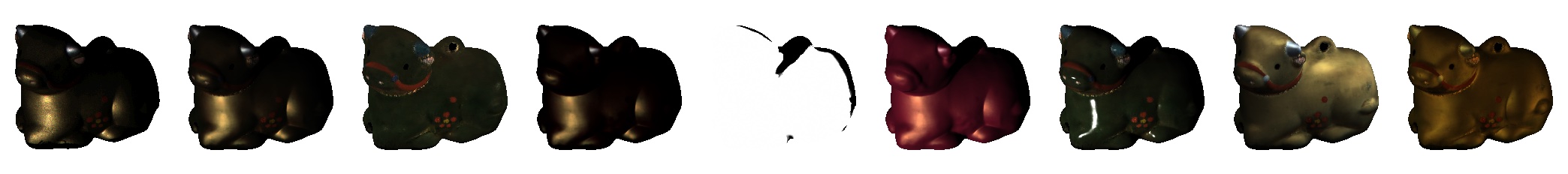}
    \includegraphics[width=\textwidth]{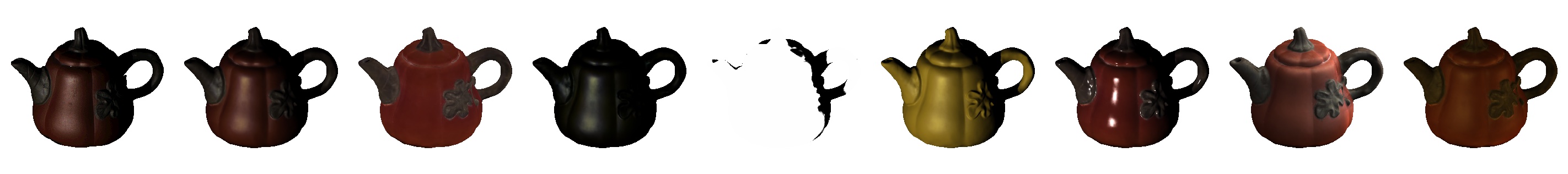}
    \includegraphics[width=\textwidth]{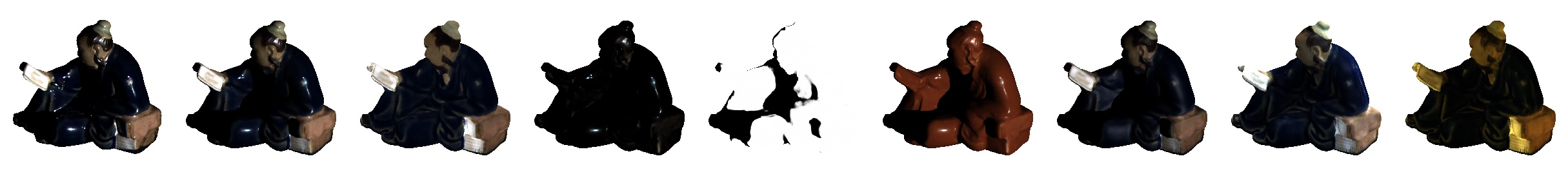}
    \caption{Scene decomposition results of our method and applications. Columns 1--2 show the ground-truth and rendered images. Columns 3--5 show the reconstructed albedo, specular component, and visibility. Columns  6--7 edit the albedo and specular component of the objects, respectively.
    Columns 8--9 are two relighting results.} \label{fig:application}
    \vspace{-3em}
\end{figure}

Our method jointly estimates surface normals, spatially-varying BRDFs, and lights. After optimization, the reconstructed object can be used for novel-view rendering, relighting, and material editing. \Fref{fig:application} shows the scene decomposition, material editing, and relighting results for a novel view of objects from \mvdiligentdata.

Please check the supplementary video for more results.

\end{document}